\documentclass[10pt,journal,compsoc]{IEEEtran}

\usepackage{ifpdf}

\ifCLASSOPTIONcompsoc
  \usepackage[nocompress]{cite}
\else
  \usepackage{cite}
\fi

\ifCLASSINFOpdf
  \usepackage[pdftex]{graphicx}
\else
\fi

\usepackage{amsmath}
\usepackage{acronym}
\usepackage{algorithmic}
\usepackage[ruled,vlined]{algorithm2e}
\usepackage{array}
\usepackage{mdwmath}
\usepackage{mdwtab}
\usepackage{eqparbox}
\ifCLASSOPTIONcompsoc
 \usepackage[caption=false,font=footnotesize,labelfont=sf,textfont=sf]{subfig}
\else
 \usepackage[caption=false,font=footnotesize]{subfig}
\fi
\usepackage{fixltx2e}

\usepackage{stfloats}
\usepackage[dvipsnames]{xcolor}

\ifCLASSOPTIONcaptionsoff
 \usepackage[nomarkers]{endfloat}
\let\MYoriglatexcaption\caption
\renewcommand{\caption}[2][\relax]{\MYoriglatexcaption[#2]{#2}}
\fi
\usepackage{url}
\usepackage{booktabs}
\usepackage{amssymb}
\usepackage{makecell}

\definecolor{collapsed}{rgb}{0.5, 0.5, 0.5}

\newcommand\MYhyperrefoptions{bookmarks=true,bookmarksnumbered=true,
pdfpagemode={UseOutlines},plainpages=false,pdfpagelabels=true,
colorlinks=true,linkcolor={black},citecolor={black},urlcolor={black},
pdftitle={Bare Demo of IEEEtran.cls for Computer Society Journals},%<!CHANGE!
pdfsubject={Typesetting},%<!CHANGE!
pdfauthor={Michael D. Shell},%<!CHANGE!
pdfkeywords={Computer Society, IEEEtran, journal, LaTeX, paper,
             template}}%<^!CHANGE!
\ifCLASSINFOpdf
\usepackage[\MYhyperrefoptions,pdftex]{hyperref}
\else
\usepackage[\MYhyperrefoptions,breaklinks=true,dvips]{hyperref}
\usepackage{breakurl}
\fi

	\newcommand{\qileft}{[\kern-0.15em[}
	\newcommand{\qiLeft}{\left[\kern-0.4em\left[}
	\newcommand{\qiright}{]\kern-0.15em]}
	\newcommand{\qiRight}{\right]\kern-0.4em\right]}

	\newcommand{\eg}{{\emph{e.g.}}}
	\newcommand{\ie}{{\emph{i.e.}}}

	\renewcommand{\Roman}[1]{\uppercase\expandafter{\romannumeral#1}}

\begin{document}
% \title{Relational Self-Supervised Learning}
\title{Weak Augmentation Guided Relational Self-Supervised Learning}

% \author{Michael~Shell,~\IEEEmembership{Member,~IEEE,}
%         John~Doe,~\IEEEmembership{Fellow,~OSA,}
%         and~Jane~Doe,~\IEEEmembership{Life~Fellow,~IEEE}% <-this % stops a space
% \IEEEcompsocitemizethanks{\IEEEcompsocthanksitem M. Shell was with the Department
% of Electrical and Computer Engineering, Georgia Institute of Technology, Atlanta,
% GA, 30332.\protect\\
% % note need leading \protect in front of \\ to get a newline within \thanks as
% % \\ is fragile and will error, could use \hfil\break instead.
% E-mail: see http://www.michaelshell.org/contact.html
% \IEEEcompsocthanksitem J. Doe and J. Doe are with Anonymous University.}% <-this % stops a space
% \thanks{Manuscript received April 19, 2005; revised August 26, 2015.}}

\author{Mingkai Zheng, 
        Shan You,
        Fei Wang, 
        Chen Qian,
        Changshui Zhang,~\IEEEmembership{Fellow,~IEEE,} \\
        Xiaogang Wang, 
        Chang Xu,~\IEEEmembership{Senior Member,~IEEE,}
\IEEEcompsocitemizethanks{
\vspace{-1mm}
\IEEEcompsocthanksitem Mingkai Zheng and Chang Xu are with the School of Computer Science, Faculty of Engineering, The University of Sydney, Australia. E-mail: mzhe4001@uni.sydney.edu.au, c.xu@sydney.edu.au \protect\\
\vspace{-1mm}
\IEEEcompsocthanksitem Shan You and Chen Qian are with the SenseTime Research. E-mail: \{youshan,qianchen\}@sensetime.com \protect\\
\vspace{-1mm}
\IEEEcompsocthanksitem Fei Wang is with the University of Science and Technology of China. E-mail: wangfei91@mail.ustc.edu.cn \protect \\
\vspace{-1mm}
\IEEEcompsocthanksitem Changshui Zhang is with the Department of Automation, Tsinghua University, Institute for Artificial Intelligence, Tsinghua University (THUAI), Beijing National Research Center for Information Science and Technology (BNRist). E-mail: zcs@mail.tsinghua.edu.cn \protect\\
\vspace{-1mm}
\IEEEcompsocthanksitem Xiaogang Wang is with Department of Electronic Engineering, The Chinese University of Hong Kong. E-mail: xgwang@ee.cuhk.edu.hk \protect\\
}% <-this % stops an unwanted space

\thanks{This work was supported in part by the Australian Research Council under Projects DP240101848 and FT230100549. (Corresponding authors: Chang Xu and Shan You.)}
% \IEEEcompsocitemizethanks{\IEEEcompsocthanksitem M. Shell was with the Department
% of Electrical and Computer Engineering, Georgia Institute of Technology, Atlanta,
% GA, 30332.\protect\\
% % note need leading \protect in front of \\ to get a newline within \thanks as
% % \\ is fragile and will error, could use \hfil\break instead.
% E-mail: see http://www.michaelshell.org/contact.html
% \IEEEcompsocthanksitem J. Doe and J. Doe are with Anonymous University.}% <-this % stops a space
% \thanks{Manuscript received April 19, 2005; revised August 26, 2015.}
}

\IEEEtitleabstractindextext{%
\begin{abstract}
Self-supervised Learning (SSL) including the mainstream contrastive learning has achieved great success in learning visual representations without data annotations. However, most methods mainly focus on the instance level information (\ie, the different augmented images of the same instance should have the same feature or cluster into the same class), but there is a lack of attention on the relationships between different instances. In this paper, we introduce a novel SSL paradigm, which we term as relational self-supervised learning  (ReSSL) framework that learns representations by modeling the relationship between different instances. Specifically, our proposed method employs sharpened distribution of pairwise similarities among different instances as \textit{relation} metric, which is thus utilized to match the feature embeddings of different augmentations. To boost the performance, we argue that weak augmentations matter to represent a more reliable relation, and leverage momentum strategy for practical efficiency. The designed asymmetric predictor head and an InfoNCE warm-up strategy enhance the robustness to hyper-parameters and benefit the resulting performance. Experimental results show that our proposed ReSSL substantially outperforms the state-of-the-art methods across different network architectures, including various lightweight networks (\eg, EfficientNet and MobileNet).
\end{abstract}

}

\maketitle

\IEEEdisplaynontitleabstractindextext
\IEEEpeerreviewmaketitle

\ifCLASSOPTIONcompsoc
\IEEEraisesectionheading{\section{Introduction}\label{sec:introduction}}
\else
\section{Introduction}
\label{sec:introduction}
\fi

\IEEEPARstart{R}{ecently}, self-supervised learning (SSL) has shown its superiority and achieved promising results for unsupervised visual representation learning in computer vision tasks \cite{cmc, deepinfomax, cpc, simclr, SimSiam, instance_discrimination, byol, moco}. The purpose of a typical self-supervised learning algorithm is to learn general visual representations from a large amount of data without human annotations, which can be transferred or leveraged in downstream tasks (\eg, classification, detection, and segmentation). Some previous works \cite{swav, byol} even have proven that a good unsupervised pretraining can lead to a better downstream performance than supervised pretraining.  

% Early self-supervised learning mainly focus on exploiting the intrinsic structural information from the image itself, and developing helpful pretext-tasks. 
Among various SSL algorithms, contrastive learning \cite{instance_discrimination, alignment_uniformity, simclr} serves as a state-of-the-art framework, which mainly focuses on learning an invariant feature from different views. For example, instance discrimination is a widely adopted pre-text task as in \cite{simclr, moco, instance_discrimination},  which utilizes the noisy contrastive estimation (NCE) to encourage two augmented views of the same image to be pulled closer on the embedding space but pushes apart all the other images away. Deep Clustering \cite{deepclustering, Self-labelling, swav} is an alternative pre-text task that forces different augmented views of the same instance to be clustered into the same class. However, instance discrimination based methods will inevitably induce a class collision problem \cite{contrastive_theory, PCL, debiased},  where similar images should be pulled closer instead of being pushed away. Deep clustering based methods cooperated with traditional clustering algorithms to assign a label for each instance, which relaxed the constraint of instance discrimination, but most of these algorithms adopt a strong assumption, \ie, the labels must induce an equipartition of the data, which might introduce some noise and hurt the learned representations.

In this paper, we introduce a novel Relational Self-Supervised Learning framework (ReSSL), which does not encourage explicitly to push away different instances, but uses \textit{relation} as a manner to investigate the inter-instance relationships and highlight the intra-instance invariance. Concretely, we aim to maintain the consistency of pairwise similarities among different instances for two different augmentations. For example, if we have three instances  $\mathbf{x}^1$, $\mathbf{x}^2$, $\mathbf{y}$ and $\mathbf{z}$ where $\mathbf{x}^1$, $\mathbf{x}^2$ are two different augmentations of $\mathbf{x}$,  $\mathbf{y}$, and $\mathbf{z}$ is a different sample. Then, if $\mathbf{x}^1$ is similar to $\mathbf{y}$ but different to $\mathbf{z}$, we wish $\mathbf{x}^2$ can maintain such relationship and vice versa. In this way, the relation can be modelled as a similarity distribution between a set of augmented images, and then use it as a metric to align the same images with different augmentations, so that the relationship between different instances could be maintained across different augmented views. 

However, this simple manner induces unexpectedly horrible performance if we follow the same training recipe as other contrastive learning methods \cite{simclr, moco}. We argue that construction of a proper relation matters for ReSSL; aggressive data augmentations as in \cite{simclr, simclrv2, goodview} are usually leveraged by default to generate diverse positive pairs that increase the difficulty of the pre-text task. However, this hurts the reliability of the target relation. Views generated by aggressive augmentations might cause the loss of semantic information, so the target relation might be noisy and not that reliable. In this way, we propose to leverage weaker augmentations to represent the relation, since much lesser disturbances provide more stable and meaningful relationships between different instances. Besides, we also sharpen the target distribution to emphasize the most important relationship and utilize the memory buffer with a momentum-updated network to reduce the demand of large batch size for more efficiency. With such simple relation metric, our ReSSL \cite{ressl} achieves 69.9\% Top-1 linear evaluation accuracy with 200 epochs ImageNet pretraining, which is 2.4\% better than our baseline (MoCo V2).

To further improve ReSSL \cite{ressl}, we leverage an asymmetric structure as in BYOL \cite{byol}, and SimSiam \cite{SimSiam} which adopts an additional prediction head on top of the projector. We show that breaking the symmetry could improve the performance and make the system robust to selecting the hyper-parameters. Moreover, we found that the target relation is not very reliable in the early stage of the training. To resolve this issue, we propose a warm-up with InfoNCE strategy, which adopts the InfoNCE from the beginning of the training process, and gradually shifts the objective to our relational metric along with the training. Note the InfoNCE objective will introduce clear positive and negative pairs, which we do not encourage in this paper. However, it can provide better guidance in the early stage of the training. In this way, the model will first perform an instance discrimination task and then gradually relax the constrain and pay more attention to maintain the relationships among the samples. With the predictor and Warm-up strategy, the linear evaluation accuracy of ReSSL can be further boosted from 69.9\% to 72.0\%. Notabaly, when working with the Multi-Crop strategy (200 epochs), ReSSL achieved new state-of-the-art performance 76.0\% Top-1 accuracy, which is 2.7\% higher than CLSA-Multi \cite{stronger}.
% this warm-up strategy, we show that the performance of ReSSL can be further boosted.

% Experimental results on multiple benchmark datasets show the superiority of ReSSL in terms of both performance and efficiency. For example, with 200 epochs of pre-training, our ReSSL achieved 72.0\% Top-1 accuracy on ImageNet \cite{imagenet_cvpr09} linear evaluation protocol, which is 4.5\% higher than our baseline method (MoCoV2 \cite{mocov2}). When working with the Multi-Crop strategy (200 epochs), ReSSL achieved new state-of-the-art performance 76.0\% Top-1 accuracy, which is 2.7\% higher than CLSA-Multi \cite{stronger}.

On the other hand, most aforementioned contrastive learning algorithms \cite{mocov2, simclr} only works for large networks (\eg ResNet50 \cite{resnet} or even larger) but has unsatisfying performance on small networks. For instance, the linear evaluation accuracy on ImageNet using MoCo V2 is only about 42.2\% with EfficientNet-B0 \cite{efficientnet}, which is 34.9\% lower than its supervised performance 77.1\%. With MobileNet-V3-Large \cite{mbv3}, the performance gaps become even larger (36.3\% vs 75.2\%). To resolve this issue, some knowledge distillation (KD) methods has been proposed to transfer the pretrained SSL features from a large network to a small network. In this paper, we show that ReSSL is very friendly to lightweight architectures since the linear evaluation performance can directly surpass previous KD algorithms (\eg SEED \cite{seed}, DisCo \cite{disco} and BINGO \cite{bingo}) without a large teacher. For example, with 200 epochs of pre-training and multi-crops strategy, ReSSL achieves 66.5\% / 71.0\% Top-1 linear evaluation accuracy with ResNet18 / ResNet34. This result is 8.6\% / 12.5\% higher than SEED, 5.9\% / 8.5\% higher than DisCo, and 4.1\% / 7.5\% better than BINGO. (All the KD methods has a ResNet-50 teacher with 67.5\% Top-1 Accuracy.)

Our contributions can be summarized as follows.
\begin{itemize}
    \item We proposed a novel SSL paradigm, which we term it as relational self-supervised learning (ReSSL). ReSSL maintains the relational consistency between the instances under different augmentations instead of explicitly pushing different instances away.
    
    \item  Our proposed weak augmentation and sharpening distribution strategy provide a stable and high quality target similarity distribution, which makes the framework works well.
    
    \item  ReSSL is a simple and effective SSL framework since it simply replaces the widely adopted contrastive loss with our proposed relational consistency loss. With a simple modification on MoCo, ReSSL achieved state-of-the-art performance with much lesser training cost. 
    
    \item ReSSL is a lightweight friendly framework. Unlike previous contrastive learning algorithms that have a poor performance on small architectures, ReSSL consistently improves the performance  for various lightweight networks and achieves better performance than KD algorithms.

\end{itemize}

\section{Related Works}
\textbf{Self-Supervised Learning with Pretext Tasks}. Self-supervised learning is a general representation learning framework that relies on the pretext tasks that can be formulated using only unlabeled data. The concept is ``Obtain labels from the data itself by using a semi-automatic process" and ``Predict part of the data from other parts" \cite{ssl_survey}. For this reason, many previous works focused on how to utilize the structural features of the images to design better pretext tasks that can learn more general representations.  Many pretext tasks were found to be conducive to learning image features. For example,  context prediction \cite{contextpred}, jigsaw puzzles \cite{jigsaw}, and rotation angle prediction \cite{rotation}. These methods can learn desired and transferable representations that achieve promising results in downstream tasks. Some of these tasks can even be used as an auxiliary task to improve the performance of the semi-supervised and supervised learning \cite{s4l, rotation_selfsup, SupervisedCL}.

\textbf{Generative Based Model}. Image generation is always one of the most popular ideas for self-supervised learning. The early generative base model aims to encode the images to a latent vector and then decode it to reconstruct the original images \cite{autoencoder, vae}. With the development of GAN \cite{gan, bigbigan}, some works \cite{biggan, bigbigan} have started to leverage the power of the discriminator to jointly differentiate both images and latent vectors. In this way, the generator not only maps latent samples to generated data but also has an inverse mapping from data to the latent representation. There are also some works aim to reconstruct the original image from the corrupted image, \eg image colorization \cite{img_color}, super-resolution \cite{srgan}, and image inpainting \cite{inpainting}.

\textbf{Instance Discrimination}.
The recent contrastive learning methods \cite{cpc, cmc, simclr, moco, goodview, pirl, huang2021self, deepinfomax, mochi} have made a lot of progress in the field of self-supervised learning. Most of the previous contrastive learning methods are based on the instance discrimination \cite{instance_discrimination} task in which positive pairs are defined as different views of the same image, while negative pairs are formed by sampling views from different images. SimCLR \cite{simclr, simclrv2} shows that image augmentation (\eg Grayscale, Random Resized Cropping, Color Jittering, and Gaussian Blur), nonlinear projection head and large batch size play a critical role in contrastive learning. Since large batch size usually requires a lot of GPU memory, which is not very friendly to most researchers. MoCo \cite{moco, mocov2} proposed a momentum contrast mechanism that forces the query encoder to learn the representation from a slowly progressing key encoder and maintain a memory buffer to store a large number of negative samples, it achieves better performance with smaller batch size.  NNCLR \cite{nnclr} reverses the purpose of the memory buffer, where the goal is to find the most similar samples from the memory buffer and treat it as the positive pair to further increase the difficulty of the pretext task. InfoMin \cite{goodview} proposed a set of stronger augmentation that reduces the mutual information between views while keeping task-relevant information intact. AlignUniform \cite{alignment_uniformity} shows that alignment and uniformity are two critical properties of contrastive learning. 

\textbf{Deep Clustering}. 
In contrast to instance discrimination which treats every instance as a distinct class, deep clustering \cite{deepclustering} adopts the traditional clustering method (\eg KMeans) to label each image iteratively. Eventually, similar samples will be clustered into the same class. Simply applying the KMeans algorithm might lead to a degenerate solution where all data points are mapped to the same cluster; SeLa \cite{Self-labelling} solved this issue by adding the constraint that the labels must induce equipartition of the data and proposed a fast version of the Sinkhorn-Knopp to achieve this. SwAV \cite{swav} further extended this idea and proposed a scalable online clustering framework by maintaining a set of prototypes and assigning the samples equally to each prototype. DINO \cite{dino} is another type of clustering framework that has a very similar objective. It replaces the Sinkhorn-Knopp with the sharpening and centering tricks, where the sharpening aims to minimize the entropy so that each sample will be assigned to a prototype, and centering aims to maximize the entropy to prevent all samples from being assigned into one prototype.

\textbf{Class Collision Problem} PCL \cite{PCL} reveals the class collision problem in contrastive learning, since instance discrimination treats every sample as a single class, the similar samples will be undesirably pushed apart by the contrastive loss. PCL solved this problem by performing the Expectation-Maximization algorithm iteratively and performing the contrastive loss and clustering loss simultaneously, although it gets the same linear classification accuracy with MoCo V2 \cite{mocov2}, it has better performance on downstream tasks. CLD \cite{cld} proposed a very similar idea where the only difference is that the clustering process will be applied on the fly in each batch. WCL \cite{WCL} proposed a novel clustering algorithm by generating the connected component from the nearest neighbor graph. Different from the clustering-based method, AdpCLR \cite{topk} and FNC \cite{FNCancel} directly find the top-K closest instances on the embedding space and treat these instances as their positive pairs to learn a much more compact representation for similar samples.

\textbf{Contrastive Learning Without Negative Samples}. 
Most previous contrastive learning methods prevent the model collapse in an explicit manner (\eg~push different instances away from each other or force different instances to be clustered into different groups.) BYOL \cite{byol} can learn a high-quality representation without negatives. Specifically, it trains an online network to predict the target network representation of the same image under a different augmented view and uses an additional predictor network on top of the online encoder to avoid model collapse. SimSiam \cite{SimSiam} shows that simple Siamese networks can learn meaningful representations even without the use of negative pairs, large batch sizes, and momentum encoders. AdCo \cite{adco} shows that a set of negative features can replace the negative samples, and these features can be adversarially learned by maximizing the contrastive loss. Barlow Twins \cite{barlowtwins} proposed a redundancy-reduction principle that aims to make the cross-correlation matrix computed from twin representations as close to the identity matrix as possible. This objective can be regarded as a dual form of CL, since CL aims to decouple the features for each instance, and Barlow Twins aims to decouple the feature dimensions.

\textbf{Contrastive Learning for Lightweight Architectures}. Although contrastive learning has achieved remarkable performance for large networks (\eg ResNet50 or larger), however, most of these methods have a horrible performance on lightweight architecture since it is pretty hard for smaller models to learn instance-level discriminative representation with a large amount of data. SEED \cite{seed} first proposed a knowledge distillation (KD) framework for contrastive learning; it requires the student network to mimic the similarity distribution inferred by the teacher over the memory buffer. DisCo \cite{disco} introduced a different KD method which aims to constrain the last embedding of the student to be consistent with that of the teacher; it also alleviates the distilling bottleNeck problem and presents a large projection head can increase the capability of the model generalization. BINGO \cite{bingo} utilized the teacher network to construct a group of similar instances; the objective is to aggregate compact representations over the student with respect
to instances in the same group.

\textbf{Relation-based Knowledge Distillation} This study is closely related to knowledge distillation, which aims to transfer knowledge from a larger teacher model to a small student model. Notably, works such as \cite{darkrank} and \cite{yu2019learning} have advanced this field by demonstrating the transfer of knowledge through the preservation of ranking orders among sample similarities. \cite{yu2019learning} enriches this approach by integrating hint and attention transfer mechanisms to maintain consistency in feature maps between the teacher and student models. In a related vein, \cite{tung2019similarity} introduces a methodology for preserving the consistency of similarity matrices derived from the activation maps between student and teacher models. Moreover, \cite{park2019relational} explores a congruent concept by ensuring the consistency of similarity distribution between the representations of different samples within a minibatch. This approach, most akin to our current work, operates within a supervised learning paradigm, leveraging a pre-trained teacher model and explicit supervisory signals. However, unlike our approach, it does not confront the challenges associated with model collapse, a critical consideration in unsupervised settings.

\label{section:method}
\section{Methodology}

In this section, we will first revisit the preliminary works on contrastive learning; then, we will introduce our proposed relational self-supervised learning framework (ReSSL). After that, the algorithm and the implementation details will also be explained.

\subsection{Preliminaries on Self-supervised Learning}
Given $N$ unlabeled samples $\mathbf{x}$, we randomly apply a composition of augmentation functions $T(\cdot)$ to obtain two different views $\mathbf{x}^{1}$ and $\mathbf{x}^{2}$ through $T(\mathbf{x}, \theta_{1})$ and $T(\mathbf{x}, \theta_{2})$ where $\theta$ is the random seed for $T$. Then, a convolutional neural network based encoder $\mathcal{F}(\cdot)$ is employed to extract the information from these samples, i.e.,   $\mathbf{h} = \mathcal{F}(T(\mathbf{x}, \theta))$. Finally, a two-layer non-linear projection head $g(\cdot)$ is utilized to map $\mathbf{h}$ into embedding space, which can be written as: $\mathbf{z} = g(\mathbf{h})$. SimCLR \cite{simclr} and MoCo \cite{moco} style framework adopt the noise contrastive estimation (NCE) objective for discriminating different instances in the dataset. Suppose $\mathbf{z}^{1}_{i}$ and $\mathbf{z}^{2}_{i}$ are the representations of two augmented views of $\mathbf{x}_{i}$ and $\mathbf{z}_{k}$ is a different instance. The NCE objective can be expressed by Eq. \eqref{equation:nce}, where $\tau$ is the temperature parameter. Noted that we take the $L_2$ normalized vectors for representation $\mathbf{z}$ by default. 
\begin{equation}
    \label{equation:nce}
    \mathcal{L}_{NCE} = -\log \frac{\exp(\mathbf{z}^{1} \cdot \mathbf{z}^{2}/ \tau) }{ \exp(\mathbf{z}_{i}^{1} \cdot \mathbf{z}_{i}^{2} / \tau ) + \sum_{k=1}^{N}  \exp(\mathbf{z}_{i}^{1} \cdot \mathbf{z}_{k} / \tau ) }.
\end{equation}

BYOL \cite{byol} and SimSiam \cite{SimSiam} style framework add an additional non-linear predictor head $q(\cdot)$ which further maps $\mathbf{z}$ to $\mathbf{p}$. The model will minimize the negative cosine similarity (equivalent to minimize the L2 distance) between $\mathbf{z}$ to $\mathbf{p}$.
\begin{align}
    \label{equation:byol}
    \mathcal{L}_{cos} &= - \frac{\mathbf{p}^1}{\lVert \mathbf{p}^1 \lVert} \cdot \frac{\mathbf{z}^2}{\lVert \mathbf{z}^2 \lVert}, & \mathcal{L}_{mse} = \lVert  \mathbf{p}^1 - \mathbf{z}^2\lVert^2_2.
\end{align}
Tricks like stop-gradient and momentum teacher are often applied to avoid model collapsing.

\begin{figure*}
    \centering
    \includegraphics[width=.8\linewidth]{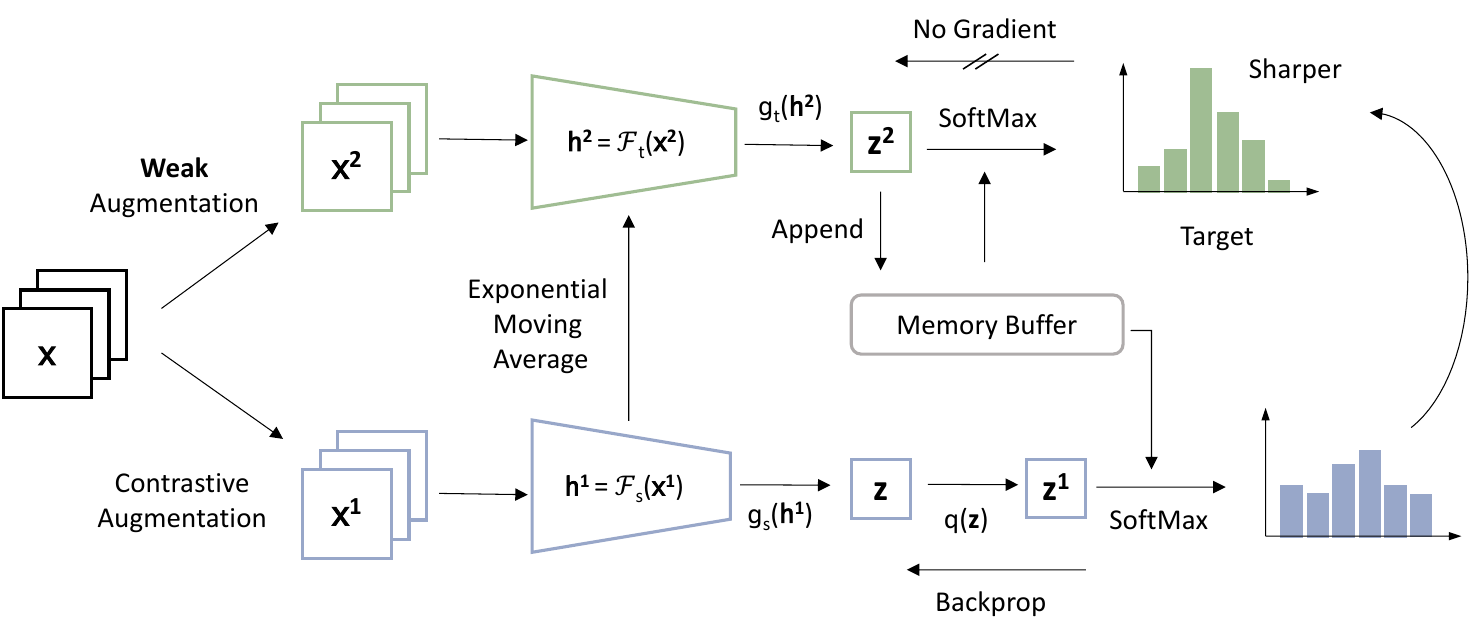}
    \vspace{-10pt}
    \caption{The overall framework of our proposed method. We adopt the student-teacher framework where the student is trained to predict the representation of the teacher, and the teacher is updated with a “momentum update” (exponential moving average) of the student. The relationship consistency is achieve by align the conditional distribution for student and teacher model. Please see more details in our method part.}
    \label{fig:framework}
    \vspace{-10pt}
\end{figure*}

\subsection{Relational Self-Supervised Learning}
In classical self-supervised learning, different instances are to be pushed away from each other, and augmented views of the same instance is expected to be of exactly the same features. However, both constrains are too restricted because of the existence of similar samples and the distorted semantic information if aggressive augmentation is adopted. In this way, we do not encourage explicit negative instances (those to be pushed away) for each instance; instead, we leverage the pairwise similarities as a manner to explore their relationships. And we pull the features of two different augmentations in this sense of relation metric. As a result, our method relaxes both \eqref{equation:nce} and \eqref{equation:byol}, where different instances do not always need to be pushed away from each other; and augmented views of the same instance only need to share the similar but not exactly the same features. 

Concretely, given a image $\mathbf{x}$ in a batch of samples , two different augmented views can be obtained by $\mathbf{x}^1 = T(\mathbf{x}, \theta_1)$, $\mathbf{x}^2 = T(\mathbf{x}, \theta_2)$ and calculate the corresponds embedding  $\mathbf{z}^1 = q(g(\mathcal{F} (\mathbf{x}^1)))$, $\mathbf{z}^2 = g(\mathcal{F}(\mathbf{x}^2))$. Then, we calculate the similarities between the instances of the first augmented images. Which can be measured by $\mathbf{z}^{1} \cdot \mathbf{z}_{i}$. A softmax layer can be adopted to process the calculated similarities, which then produces a relationship distribution:
\begin{align}
    \mathbf{p}^1_i &= \frac{\exp(\mathbf{z}^{1} \cdot \mathbf{z}_{i}/ \tau_s) }{\sum_{k=1}^{K}  \exp(\mathbf{z}^{1} \cdot \mathbf{z}_{k} / \tau_s) }. \label{equation:teacher}
\end{align}
where $\tau_s$ is the temperature parameter. At the same time, we can calculate the relationship between $\mathbf{x}^2$ and the $i$-th instance as $\mathbf{z}^{2} \cdot \mathbf{z}_{i}$. The resulting relationship distribution can be written as:
\begin{align}
    \mathbf{p}^2_i &= \frac{\exp(\mathbf{z}^{2} \cdot \mathbf{z}_{i} / \tau_t) }{\sum_{k=1}^{K}  \exp(\mathbf{z}^{2} \cdot \mathbf{z}_{k} / \tau_t) }. \label{equation:student}
\end{align}
where $\tau_t$ is a different temperature parameter. We propose to push the relational consistency between $p^{1}_i$ and $p^{2}_i$ by minimizing the Kullback–Leibler divergence, which can be formulated as:
\begin{equation}
    \label{equation:loss}
    \mathcal{L}_{relation} =  D_{KL} (\mathbf{p}^2 || \mathbf{p}^1) = H(\mathbf{p}^2, \mathbf{p}^1) - H(\mathbf{p}^2).
\end{equation}
Since the $H(\mathbf{p}^2)$ will be a constant value, we only minimize $H(\mathbf{p}^2,\mathbf{p}^1)$ in our implementation.

\textbf{More efficiency with Momentum targets.} 
However, the quality of the target similarity distribution $\mathbf{p}^2$ is crucial, to make the similarity distribution reliable and stable, we usually require a large batch size which is very unfriendly to GPU memories. To resolve this issue, we utilize a ``momentum update" network as in \cite{moco, mocov2}, and maintain a large memory buffer $\mathcal{Q}$ of $K$ past samples $\{\mathbf{z}_k | k = 1, ... , K \}$ (following the FIFO principle) for storing the feature embeddings from the past batches, which can then be used for simulating the large batch size relationship and providing a stable similarity distribution. 
\begin{equation} \label{equation:ema}
    \mathcal{F}_t \leftarrow m \mathcal{F}_t + (1 - m)\mathcal{F}_s, \quad  g_t \leftarrow m g_t + (1 - m) g_s,
\end{equation}
where $\mathcal{F}_s$ and $g_s$ denote the most latest encoder and head, respectively, so we name them as the student model with a subscript $s$. On the other hand, $\mathcal{F}_t$ and $g_t$ stand for ensembles of the past encoder and head, respectively, so we name them as the teacher model with a subscript $t$. $m$ represents the momentum coefficient which controls how fast the teacher $\mathcal{F}_t$ will be updated.

\textbf{Sharper Distribution as Target}.
 Note, the value of $\tau_t$ has to be smaller than $\tau_s$ since $\tau_t$ will be used to generate the target distribution. A smaller $\tau$ will result in a ``sharper" distribution which can be interpreted as highlighting the most similar feature for $\mathbf{z}^2$, since the similar features will have a higher value and the dissimilar features will have a lower value. Align $\mathbf{p}^1$ with $\mathbf{p}^2$ can be regarded as pulling $\mathbf{z}^1$ towards the features that are similar with $\mathbf{z}^2$. In this way, ReSSL makes the similar samples more similar, and dissimilar ones be more dissimilar.

\textbf{Weak Augmentation Strategy for Teacher}.
To further improve the quality and stability of the target distribution, we adopt a weak augmentation strategy for the teacher model since the standard contrastive augmentation is too aggressive, which introduced too many disturbances and will mislead the student network. Please refer to more details in our empirical study.

\textbf{Compare with SEED and CLSA}.  SEED \cite{seed} follows the standard Knowledge Distillation (KD) paradigm \cite{hinton2015distilling,you2017learning,du2020agree} where it aims to distill the knowledge from a larger network into a smaller architecture. The knowledge transfer happens in the same view but between different models. In our framework, we are trying to maintain the relational consistency between different augmentations; the knowledge transfer happens between different views but in the same network. CLSA \cite{stronger} also introduced the concept of using weak augmentation to guide a stronger augmentation. However, the ``weak" augmentation in CLSA is equivalent to the ``strong" augmentation in our method (We do not use any stronger augmentations such as \cite{autoaugment, randaugment} in our default setting). On the other hand, CLSA requires at least one additional sample during training, which requires a lot more GPU memories and slow down the training speed.

\subsection{Improving ReSSL with Predictor and InfoNCE}
\textbf{Asymmetric Head for Student}. For the student network, we adopt an additional predictor head as in \cite{byol,SimSiam}, which breaks the symmetry of the Siamese network. \cite{SimSiam} demonstrates that the typical contrastive learning frameworks are just an Expectation-Maximization like algorithm; the predictor head helps fill the gaps between an augmented view and the expectation. In ReSSL,  Breaking the symmetry could make the system become robuster to the selection of the temperature, and also slightly improve the performance. Please see more empirical analysis in Section 4. 

\textbf{Warm-up with InfoNCE}. As we have described that the sharpening strategy is a crucial point in ReSSL, it can be interpreted as pulling $\mathbf{z}^{1}$ towards to features that are similar with $\mathbf{z}^{2}$. However, the network is not able to capture high-quality similar samples especially in the early stage of the training, which makes the target distribution $\mathbf{p}^{2}$ unreliable. To resolve this issue,
we add an additional InfoNCE loss to cooperate with our relational consistency loss, which can be expressed by Eq \eqref{equation:warmup}
\begin{equation}
    \label{equation:warmup}
    \mathcal{L}_{total} =  \alpha \mathcal{L}_{relation} + (1 - \alpha)  \mathcal{L}_{InfoNCE}
\end{equation}
Where $\alpha$ is the hyper-parameter to control the weight of the two losses. Here, we set $\alpha = current\_step / warmup\_step$. In this case, $\alpha$ will increase from $0$ to $1$ along our optimization step. Basically, $\mathbf{z}^{1}$ will be pulled towards to $\mathbf{z}^{2}$ in the early stage since $\mathbf{p}^{2}$ is not reliable, then, the objective function will gradually shit from Eq. \eqref{equation:nce}  to Eq. \eqref{equation:loss}. Although InfoNCE still introduces the clear positives and negatives, the weight will decrease along with the training. Experimental results show that warm-up with InfoNCE significantly improves the results of the linear evaluation. Please see more details in Section 4 below.

% \begin{algorithm}
% \SetAlgoLined
% \SetKwInOut{Input}{Input}
% \Input{
% $\mathbf{x} $: a batch of samples. $T_w(\cdot)$: Weak augmentation function. $T_c(\cdot)$: Contrastive augmentation function. $\mathcal{F}_t$ and $\mathcal{F}_s$: the teacher and student backbone network. $g_t$ and $g_s$ : the non-linear projection head for teacher and student. $q$: the predictor head. $Q$: the memory buffer  }
% \While{network not converge} {
%     \For{i=1 to step}{
%         Fetch $\mathbf{x}$ from current batch $\mathcal{B}$
        
%         $\mathbf{z}^1 = q(g_s( \mathcal{F}_s(T_c(\mathbf{x}, \theta_1)) ))$
        
%         $\mathbf{z}^2 = g_t( \mathcal{F}_t(T_w(\mathbf{x}, \theta_2)) )$
        
%         $\mathbf{p}^1$ = SoftMax($\mathbf{z}^1 Q^T$ / $\tau_s$ ) \tcp*{ Eq. \eqref{equation:teacher}}
        
%         $\mathbf{p}^2$ = SoftMax($\mathbf{z}^2 Q^T$ / $\tau_t$ )  \tcp*{ Eq. \eqref{equation:student}}
        
%         Calculate $\mathcal{L}_{relation}$ loss \tcp*{ Eq. \eqref{equation:loss}}
        
%         Update $\mathcal{F}_s$ and $g_s$ with loss $\mathcal{L}_{relation}$
        
%         Update $\mathcal{F}_t$  by $\mathcal{F}_t \leftarrow m \mathcal{F}_t + (1 - m) \mathcal{F}_s$
        
%         Update $g_t$ by $g_t \leftarrow m g_t + (1 - m) g_s$
        
%         Update the memory buffer $Q$ by $\mathbf{z}^2$
%     }
% }
% \SetKwInOut{Output}{Output}
% \Output{The well trained model $\mathcal{F}_s$}
% \caption{Relational Self-supervised Learning}
% \label{alg:overall}
% \end{algorithm}

\section{Empirical Study of ReSSL}
In this section, we will empirically study our ReSSL on 4 popular self-supervised learning benchmarks. Specifically, we ablate the effectiveness of the sharpening, weak augmentation, predictor, and the InfoNCE warm-up strategy.

\textbf{Small Dataset}.
CIFAR-10 and CIFAR-100 \cite{cifar}. The CIFAR-10 dataset consists of 60000 32x32 colour images in 10 classes, with 6000 images per class. There are 50000 training images and 10000 test images. CIFAR-100 is just like the CIFAR-10, except it has 100 classes containing 600 images each. There are 500 training images and 100 testing images per class.

\textbf{Medium Dataset}.
STL-10 \cite{stl10} and Tiny ImageNet \cite{tinyImagenet}. STL10 \cite{stl10} dataset is composed of 96x96 resolution images with 10 classes. It contains 5K labeled training images, 8K validation images, and 100K unlabeled images. The Tiny ImageNet dataset is composed of 64x64 resolution images with 200 classes, which has 100K training images and 10k validation images.

% \renewcommand\arraystretch{1.0}
% \begin{table}[t]
%  \centering
%  \setlength\tabcolsep{4pt}
%  \caption{Compare to our baseline method on small and medium dataset. We report the results for with and without predictors}
%  \vspace{-10pt}
%  \label{table:compare_smalldata}
% \begin{tabular}{l | c c c c} 
% Method & CIFAR-10 & CIFAR-100 & STL-10 & Tiny ImageNet \\ \hline
% MoCo V2  \cite{mocov2} & 86.16 & 59.51 & 85.88 & 43.36 \\  \hline
% ReSSL         & 90.20 & 63.79 & 88.25 & 46.60  \\ 
% + Predictor          & \textbf{90.76} & \textbf{65.03} & \textbf{88.89} & \textbf{46.64} \\
% + InfoNCE Warmup          & \textbf{90.76} & \textbf{65.03} & \textbf{88.89} &\textbf{46.64}  \\
% \end{tabular}
% \vspace{-15pt}
% \end{table}

\renewcommand\arraystretch{1.0}
\begin{table*}[h]
 \centering
 \setlength\tabcolsep{3pt}
 \caption{Effect of different $\tau_t$ and $\tau_s$ for ReSSL \textbf{with} predictor}
 \vspace{-10pt}
 \label{table:ablation_t_pred}
\begin{tabular}{l c c c c c c c c c c c} 
\toprule 
Dataset        & $\tau_s$ & $\tau_t = 0.01$ & $\tau_t = 0.02$ & $\tau_t = 0.03$ & $\tau_t = 0.04$ & $\tau_t = 0.05$ & $\tau_t = 0.06$ & $\tau_t = 0.07$ & $\tau_t = 0.08$  & $\tau_t = 0.09$  & $\tau_t = 0.10$ \\  \hline
CIFAR-10       & 0.1 & 89.72  & 90.16 & 90.62 & \textbf{90.76}  & 90.46 & 90.21 & 89.95 & 89.67 & 89.61 & 89.25 \\
CIFAR-10       & 0.2 & 90.18  & 90.42 & 90.42 & 89.91 & 89.97 & 90.00 & 89.84 & 89.19 & 88.88 & 88.63 \\ \hline
CIFAR-100      & 0.1 & 63.63  & 64.61 & \textbf{65.03} & 64.60 & 64.26 & 63.19 & 62.54 & 60.56 & 60.17 & 58.18 \\
CIFAR-100      & 0.2 & 63.12  & 63.09 & 62.08 & 61.93 & 61.20 & 60.48 & 59.25 & 58.52 & 58.11 & 56.36 \\ \hline
STL-10         & 0.1 & 87.82  & 88.54 & \textbf{88.89} & 88.41 & 87.60 & 87.24 & 86.87 & 84.04 & 83.67 & 83.59 \\
STL-10         & 0.2 & 87.66  & 87.28 & 86.51 & 86.19 & 85.75 & 85.75 & 85.58 & 85.61 & 85.37 & 84.52 \\ \hline
Tiny ImageNet  & 0.1 & 45.66  & 45.62 & \textbf{46.64} & 46.06 & 45.36 & 41.22 & 37.02 & 36.03 & 32.54 & 32.28 \\
Tiny ImageNet  & 0.2 & 45.64  & 45.74 & 43.72 & 42.62 & 42.02 & 40.72 & 38.24 & 34.74 & 34.12 & 30.10 \\
\bottomrule 
\end{tabular}
\vspace{-10pt}
\end{table*}

\renewcommand\arraystretch{1.0}
\begin{table*}[h]
 \centering
 \setlength\tabcolsep{3pt}
 \caption{Effect of different $\tau_t$ and $\tau_s$ for ReSSL \textbf{without} predictor.}
 \vspace{-10pt}
 \label{table:ablation_t}
\begin{tabular}{l c c c c c c c c c c c} 
\toprule 
Dataset        & $\tau_s$ & $\tau_t = 0.01$ & $\tau_t = 0.02$ & $\tau_t = 0.03$ & $\tau_t = 0.04$ & $\tau_t = 0.05$ & $\tau_t = 0.06$ & $\tau_t = 0.07$ & $\tau_t = 0.08$  & $\tau_t = 0.09$  & $\tau_t = 0.10$ \\  \hline
CIFAR-10       & 0.1 & 89.35  & 89.74 & 90.09 & 90.04 & \textbf{90.20} & 90.18 & 88.67 &  \textcolor{collapsed}{10.00} &  \textcolor{collapsed}{10.00} & \textcolor{collapsed}{10.10} \\
CIFAR-10       & 0.2 & 89.52  & 89.67 & 89.24 & 89.50 & 89.22 & 89.40 & 89.50 & 89.58 & 89.43 & 89.43 \\ \hline
CIFAR-100      & 0.1 & 62.34  & 62.79 & 62.71 & \textbf{63.79} & 63.46 & 63.20 & 61.31 &  \textcolor{collapsed}{1.00} &  \textcolor{collapsed}{1.00} &  \textcolor{collapsed}{1.00} \\
CIFAR-100      & 0.2 & 60.37  & 60.05 & 60.24 & 60.09 & 59.09 & 59.12 & 59.76 & 59.97 & 59.69 & 59.08 \\ \hline
STL-10         & 0.1 & 86.65  & 86.96 & 87.16 & 87.32 & \textbf{88.25} & 87.83 & 87.08 & \textcolor{collapsed}{10.00} &  \textcolor{collapsed}{10.00} &  \textcolor{collapsed}{10.00} \\
STL-10         & 0.2 & 85.17  & 86.12 & 85.01 & 85.67 & 85.21 & 85.51 & 85.28 & 85.93 & 85.56 & 85.58 \\ \hline
Tiny ImageNet  & 0.1 & 45.20  & 45.40 & 46.30 & \textbf{46.60} & 45.08 & 45.24 & 44.18 &  \textcolor{collapsed}{0.50} &  \textcolor{collapsed}{0.50} &  \textcolor{collapsed}{0.50} \\
Tiny ImageNet  & 0.2 & 43.28  & 42.98 & 43.58 & 42.12 & 42.70 & 42.76 & 42.60 & 41.46 & 41.08 & 40.70 \\
\bottomrule 
\end{tabular}
\vspace{-10pt}
\end{table*}

\textbf{Implementation Details}
We adopt the ResNet18 \cite{resnet} as our backbone network. Because most of our dataset contains low-resolution images, we replace the first 7x7 Conv of stride 2 with 3x3 Conv of stride 1 and remove the first max pooling operation for a small dataset. For data augmentations, we use the random resized crops (the lower bound of random crop ratio is set to 0.2), color distortion (strength=0.5) with a probability of 0.8, and Gaussian blur with a probability of 0.5. The images from the small and medium datasets will be resized to 32x32 and 64x64 resolution respectively. Our method is based on MoCoV2 \cite{mocov2}; we use the global BN and shuffle BN  for with and without predictor setting. The momentum value and memory buffer size are set to 0.99 and 4096/16384 for small and medium datasets respectively. Moreover, The model is trained using SGD optimizer with a momentum of 0.9 and weight decay of $5e^{-4}$. We linear warm up the learning rate for 10 epochs until it reaches $0.06 \times BatchSize / 256$, then switch to the cosine decay scheduler  \cite{cosine_lr}.

\textbf{Evaluation Protocol}.
All the models will be trained for 200 epochs. For testing the representation quality, we evaluate the pre-trained model on the widely adopted linear evaluation protocol - We will freeze the encoder parameters and train a linear classifier on top of the average pooling features for 100 epochs. To test the classifier, we use the center crop of the test set and computes accuracy according to predicted output. We train the classifier with a learning rate of 10, no weight decay, and momentum of 0.9. The learning rate will be times 0.1 in 60 and 80 epochs. Note, for STL-10; the pretraining will be applied on both labeled and unlabeled images. During the linear evaluation, only the labeled 5K images will be used.

% \textbf{Result}.
% As we can see the result in Table \ref{table:compare_smalldata}, our proposed method outperforms MoCo v2 on all four benchmarks with a large margin. We can also observe that ReSSL has a slightly better performance when working with predictor.

\subsection{A Properly Sharpened Relation is A Better Target}
The temperature parameter is very crucial in most contrastive learning algorithms. To verify the effective of $\tau_s$ and $\tau_t$ for our proposed method, we fixed $\tau_s = 0.1$ or $0.2$, and sweep over $\tau_t = \{0.01, 0.02, ..., 0.10\}$. The result is shown in Table \ref{table:ablation_t_pred}. Note, here we show the result for both with and without predictor setting. For $\tau_t$, the optimal value is $0.03$ and $0.04$ across all different datasets. As we can see, the performance is increasing when we increase $\tau_t$ from $0$. After the optimal value, the performance will start to decrease.

Note, $\tau_t \rightarrow 0$ correspond to the Top-1 or $argmax$ operation, which produce a one-hot distribution as the target; this actually makes our relational consistent loss degrades to a standard InfoNCE objective where the positive pair is the most similar sample from the memory buffer instead of the augmentation of the original image. In this case, the system will be pretty similar with NNCLR \cite{nnclr}, the only difference is that NNCLR does not adopt the weak augmentation strategy, and the negative samples come from the current batch instead of the memory buffer.  On the other hand, when $\tau_t \rightarrow 0.1$, the target will be a much flatter distribution that cannot highlight the most similar features for students. Hence, $\tau_t$ can not be either too small or too large, but it has to be smaller than $\tau_s$ (\ie $\mathbf{p}^2$ has to be sharper than $\mathbf{p}^1$), therefore the target distribution can provide effective guidance to the student network. 

% On the other hand, when $\tau_t \rightarrow 0.1$, our ReSSL will collapse without the predictor since the network can simply output a constant vector to minimize our loss function. The most significant benefit of the predictor is the asymmetric system which prevents the model from collapsing. Although ReSSL also works well without the predictor, we would like to keep it since it makes the system more robust to the temperature selection. Moreover, increasing $\tau_t$  will make the target be a much flatter distribution that cannot highlight the most similar features for students. Hence, $\tau_t$ can not be either too small or too large, but it has to be smaller than $\tau_s$ ($\mathbf{p}^1$ has to be sharper than $\mathbf{p}^2$) so that the target distribution can provide effective guidance to the student model. 

For $\tau_t$, it is clearly to see that the result of $\tau_s = 0.1$ can always result a much higher performance than $\tau_s = 0.2$, which is different to MoCoV2 where $\tau_s = 0.2$ is the optimal value. According to \cite{NormFace, CosFace, ArcFace}, a greater temperature will result in a larger angular margin in the hypersphere. Since MoCoV2 adopts instance discrimination as the pretext task, a large temperature can enhance the compactness for the same instance and discrepancy for different instances. In contrast to instance discrimination, our method can be interpreted as pulling similar instances closer on the hypersphere; the similar instances might not be very reliable, especially in the early stage of the training process. Thus, the large angular margin might hurt the performance when the ground truth label is not available. 

\renewcommand\arraystretch{1.0}
\begin{table*}
 \centering
 \caption{Effect of different augmentation for teacher model}
 \vspace{-10pt}
 \label{table:augmentation}
\begin{tabular}{c c c c c c c c c} 
\toprule 
Random Resized Crops & Random Flip & Color Jitter & GrayScale & Gaussian Blur & CIFAR-10 & CIFAR-100 & STL-10 & Tiny ImageNet \\ \hline
 &   &  &  &  & 79.33 & 52.77 & 86.01 & 36.76 \\ \hline 
 \checkmark   &   &  &  &  & 90.62 & 64.98 & 88.62 & 46.38 \\ 
 & \checkmark &   &  &  & 80.38 & 51.60 & 86.24 & 38.20 \\
 & & \checkmark &  &  & 78.47 & 51.73 & 85.92 & 36.64 \\
 & & & \checkmark &   & 78.44 & 49.76 & 87.12 & 36.32 \\
 & & & & \checkmark & 78.54 & 50.65 & 85.34 & 36.20 \\ \hline
 \checkmark   & \checkmark  &  &  &  & \textbf{90.76} & \textbf{65.03} & \textbf{88.89} & \textbf{46.64} \\ \hline
 \checkmark   &   & \checkmark  &  &  & 89.80 & 62.87 & 87.45 & 43.98 \\
 \checkmark   &   &  & \checkmark &  & 89.56 & 62.61 & 88.65 & 42.52 \\
 \checkmark   &   &  &  & \checkmark & 89.18 & 62.75& 87.34 & 46.18 \\
 \checkmark   & \checkmark  & \checkmark  &  &  & 90.15 & 63.55 & 87.21 & 44.00 \\
 \checkmark   & \checkmark  &  & \checkmark &  & 89.63 & 62.95 & 88.47 & 41.88 \\
 \checkmark   & \checkmark  &  &  & \checkmark & 89.43 & 62.37 & 87.22 & 45.30 \\
 \checkmark   & \checkmark  & \checkmark & \checkmark & \checkmark & 86.88 & 58.60 & 85.05 & 38.66 \\
\toprule 
\end{tabular}
\vspace{-10pt}
\end{table*}

% \vspace{-10pt}
\subsection{Asymmetric Structure Makes Robuster System} 
Our default setting adopts the predictor structure as in \cite{byol, SimSiam} which breaks the symmetry of the system. It is however interesting to evaluate the performance with the typically symmetric system by removing the predictor structure from ReSSL. Thus, we follow the same experiment setting and present the result in Table \ref{table:ablation_t}. As we can observe that in the case of $\tau_s = 0.1$ and $\tau_t \ge 0.08$ (the gray numbers), the symmetric system will collapse since the network can simply output a constant vector to minimize our loss function. Although ReSSL also works well without the predictor, we would like to keep it since it makes the system more robust to the temperature selection. We can also notice that the performance will be slightly better with the predictor in various temperature settings.

\subsection{Working with InfoNCE Warm-up} 
Although a stand-alone relation metric already works well, the obvious problem is that the teacher network cannot capture high-quality relationships among different instances in the early stage of the training process. We addressed this issue by including the InfoNCE during the warm-up stage, as described in Section 3.3. We show the result in Table \ref{table:ablation_wamrup} below. As can be seen, the warm-up strategy significantly improves the performance on CIFAR-100, STL-10, and TinyImageNet.

\renewcommand\arraystretch{1.0}
\begin{table}[h]
 \centering
 \vspace{-5pt}
 \caption{Working with InfoNCE Warm-up}
 \vspace{-10pt}
 \label{table:ablation_wamrup}
\begin{tabular}{l c c } 
\toprule 
Dataset & w/o Warm-up & w/ Warm-up  \\ \hline
CIFAR-10 & \textbf{90.76} & 89.98   \\ \hline
CIFAR-100 & 65.03 & \textbf{65.33}   \\ \hline
STL-10 & 88.89 & \textbf{90.00}   \\ \hline
Tiny ImageNet & 46.64 & \textbf{48.22}   \\
\bottomrule 
\end{tabular}
\vspace{-10pt}
\end{table}

\subsection{Weak Augmentation Makes Better Relation}
\begin{figure}
    \centering
    \includegraphics[width=1\linewidth]{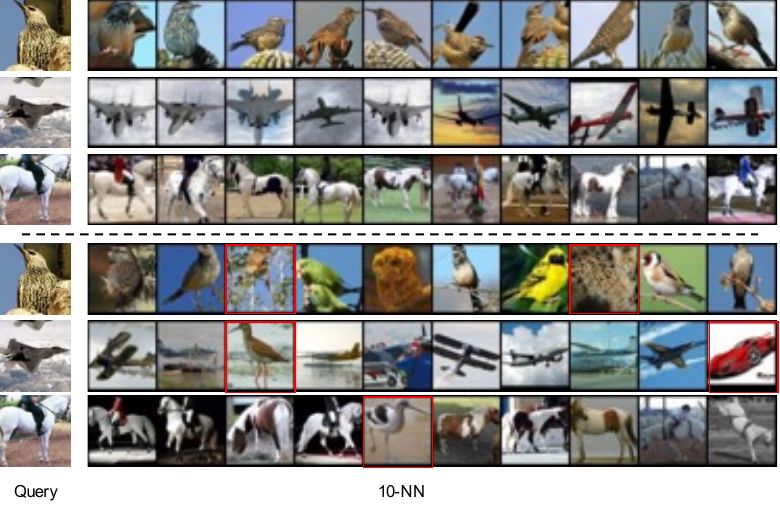}
    \vspace{-20pt}
    \caption{Visualization of the 10 nearest neighbour of the query image. The top half is the result when we apply the weak augmentation. The bottom half is the case when the typical contrastive augmentation is adopted. Note, we use the red square box to highlight the images that has different ground truth label with the query image.}
    \label{fig:nn}
    \vspace{-12pt}
\end{figure}

As we have mentioned, the weaker augmentation strategy for the teacher model is the key to the success of our framework. Here, We implement the weak augmentation as a random resized crop (the random ratio is set to $(0.2, 1)$) and a random horizontal flip.  For temperature parameter, we simply adopt the same setting as in Table \ref{table:ablation_t_pred} and report the performance of the best setting. The result is shown in Table \ref{table:ablation_weakaug}, as we can see that when we use the weak augmentation for the teacher model, the performance is significantly boosted across all datasets. We believe that this phenomenon is because relatively small disturbances in the teacher model can provide more accurate similarity guidance to the student model. To further verify this hypothesis, we random sampled three image from STL-10 training set as the query images, and then find the 10 nearest neighbour based on the weak / contrastive augmented query. We visualized the result in Figure \ref{fig:nn}, 

\renewcommand\arraystretch{1.0}
\begin{table}[h]
 \centering
 \caption{Effect of weak augmentation guided ReSSL }
 \vspace{-10pt}
 \label{table:ablation_weakaug}
\begin{tabular}{l c c } 
\toprule 
Dataset & Contrastive Aug & Weak Aug  \\ \hline
CIFAR-10 & 86.88 & \textbf{90.76}   \\ \hline
CIFAR-100 & 58.60 & \textbf{65.03}   \\ \hline
STL-10 & 85.05 & \textbf{88.89}   \\ \hline
Tiny ImageNet & 38.66 & \textbf{46.64}   \\
\bottomrule 
\end{tabular}
\vspace{-10pt}
\end{table}

\subsection{More Experiments on Weak Augmentation}
Since the weak augmentation for the teacher model is one of the crucial points in ReSSL, we further analyze the effect of applying different augmentations on the teacher model. In this experiment, we simply adopt the best temperature setting from Table \ref{table:ablation_t_pred} for each datasets and report the linear evaluation performance across the four benchmark dataset.  The results are shown in Table \ref{table:augmentation}. The first row is the baseline, where we simply resize all images to the same resolution (no extra augmentation is applied). Then, we applied random resized crops, random flip, color jitter, grayscale, gaussian blur, and various combinations. We empirically find that if we use no augmentation (\eg, no random resized crops) for the teacher model, the performance tends to degrade. This might result from that the gap of features between two views is way too smaller, which undermines the learning of representations. However, too strong augmentations of teacher model will introduce too much noise and make the target distribution inaccurate (see Figure \ref{fig:nn}). Thus mildly weak augmentations are better option for the teacher, and random resized crops with random flip is the combination with the highest performance as Table \ref{table:augmentation} shows.

\begin{figure*}[h]
    \vspace{-10pt}
    \centering
    \subfloat[ReSSL w/ Contrastive Aug]{\label{fig:moco_tsne}\includegraphics[width=0.25\linewidth]{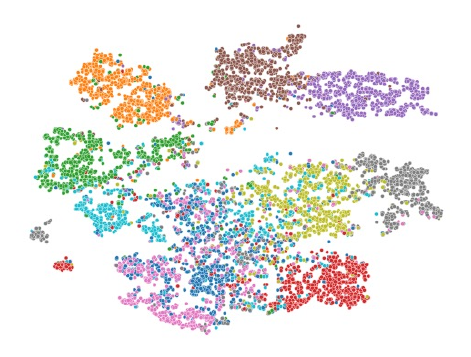}}
    \subfloat[ReSSL w/ Weak Aug]{\label{fig:moco_tsne}\includegraphics[width=0.25\linewidth]{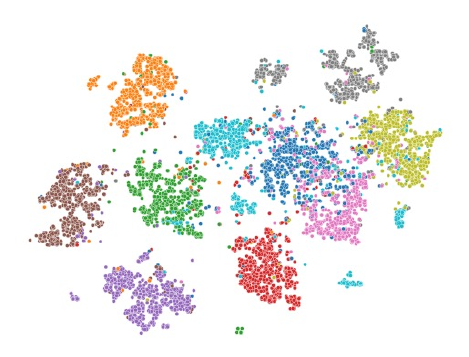}}
    \subfloat[MoCo V2]{\label{fig:moco_tsne}\includegraphics[width=0.25\linewidth]{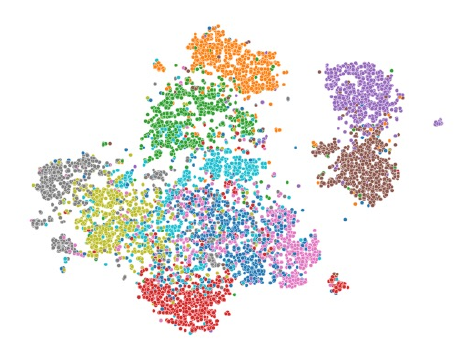}}
    \caption{t-SNE visualizations on CIFAR-10. Classes are indicated by colors. Here we show the visualization result for ReSSL with contrastive augmentation, Standard ReSSL with weak augmentation, and MoCo V2. }
    \vspace{-10pt}
    \label{fig:tsne}
\end{figure*}

\subsection{Dimension of the Relation}
Since we also adopt the memory buffer as in MoCo \cite{moco}, the buffer size will be equivalent to the dimension of the distribution $\mathbf{p}^1$ $\mathbf{p}^2$. Thus, it will be one of the crucial points in our framework. To verify the effect of the memory buffer size, we simply adopt the best temperature setting from Table \ref{table:ablation_t_pred} for each dataset, then varying the memory buffer size from 256 to 32768. The result is shown in Table \ref{table:ablation_buffer}, as we can see that the performance could be significantly improved when we increase the buffer size from 256 to 8192. However, a further increase in the buffer size (\ie 16384) can only bring a marginal improvement when the buffer is large enough. We can also observe that the performance will be slightly worse when $K=32768$. This is possibly due to a too large memory buffer involving a lot of stale embeddings, which hurts the performance.
\renewcommand\arraystretch{1.0}
\begin{table}[h]
 \centering
 \caption{Effect of different memory buffer size on small and medium dataset}
 \vspace{-10pt}
 \label{table:ablation_buffer}
\begin{tabular}{l c c c c} 
\toprule 
$K$ & CIFAR-10 & CIFAR-100 & STL-10 & Tiny ImageNet \\ \hline
256   & 89.52 & 62.21 & 85.05  & 42.80 \\
512   & 90.38 & 63.06 & 85.76  & 43.40 \\
1024  & 90.37 & 64.32 & 87.10  & 44.72 \\
4096  & 90.76 & 65.03 & 87.90  & 45.47 \\
8192  & 90.73 & \textbf{65.65} & 87.71  & 46.34 \\
16384 & \textbf{90.82}  & 65.62 & \textbf{88.89}  & \textbf{46.64} \\
32768 & 90.27 & 65.37 & 88.60  & 45.95 \\
\bottomrule 
\end{tabular}
\vspace{-10pt}
\end{table}

\subsection{Visualization of Learned Representations}
We also show the t-SNE \cite{tsne} visualizations of the representations learned by ReSSL with contrastive augmentation, ReSSL with weak augmentation, and MoCo V2 on the test set of CIFAR-10. Obviously, ReSSL with weak augmentation leads to better class separation than the contrastive loss.

\section{Experiments}

\subsection{Results on Large-scale Datasets}
We also performed our algorithm on the large-scale ImageNet-1k dataset \cite{imagenet_cvpr09}. In the experiments, we use a batch size of 1024, and adopt the LARS \cite{lars} optimizer with learning rate equals to $0.6 * BatchSize / 256$, momentum 0.9, and weight decay $1e^{-6}$. The optimizer will be scheduled by a cosine learning rate scheduler \cite{cosine_lr} with 10 epochs of Warm-up. We also adopt a memory buffer to save 65536 past examples. The momentum coefficient is $m=0.996$ and increases to 1 with a cosine schedule. For the projector, we use a 2-layer non-linear MLP network with BN and ReLU; the hidden dimension is 4096, and the output dimension is 256. The predictor consists of the same structure as the projector, but input dimension is 256.  For $\tau_t$ and $\tau_s$, we simply adopt the best setting from Table \ref{table:ablation_t_pred} where $\tau_t = 0.03$ and $\tau_s = 0.1$. For contrastive augmentation, we take the strategy from \cite{byol} with a few modifications on the parameters. Specifically, we set the probability of blurring / solarization to 50\% / 10\%, and also change the minimum crop area to 14\%. For weak augmentation, we follow the best setting in Table \ref{table:augmentation}.

\textbf{Linear Evaluation}.
For the linear evaluation of ImageNet-1k, we follow the same setting in SimCLR \cite{swav}.  The results are shown in Table \ref{table:200epoch}. Specifically, we train the linear classification layer for 90 epochs using a SGD optimizer with a Nesterov momentum value of 0.9, weight decay of 0, learning rate with $0.1 * BatchSize / 256$, and a cosine-annealed learning rate scheduler. The result is shown in Table \ref{table:200epoch}. For a fair comparison, we divide previous methods into 1x backprop and 2x backprop settings since 2x backprop methods generally require more GPU memories and longer training. As we can see clearly that ReSSL consistently outperforms previous methods on both setting. For 1x setting, ReSSL surpass previous state-of-the-art (InfoMin) 1.9\%. For 2x setting, ReSSL has 0.4\% improvements over MoCo v3 with only 2/3 training costs and smaller batch size. We also report the performance when working without the predictor and InfoNCE warm-up. In this case, the performance drops 2.1\% and 1.8\% for 1x and 2x setting, respectively, which further verified the effectiveness of these two strategies.

\renewcommand\arraystretch{1.0}
\begin{table}[h] 
 \setlength\tabcolsep{3pt}
 \centering
 \vspace{-8pt}
 \caption{Top-1 accuracy under the linear evaluation on ImageNet with the ResNet-50 backbone. ReSSL with * denote the results without predictor and InfoNCE warm-up.}
 \label{table:200epoch}
 \vspace{-10pt}
\begin{tabular}{l c c c c } 
\toprule 
Method &  Backprop  & Batch Size  & Epochs & Top-1 \\
Supervised & 1x & 256 &  120 & 76.5 \\ \hline
\emph{1x Backprop Methods } \\
NPID \cite{instance_discrimination}      & 1x &  256 &  200 & 58.5 \\
LocalAgg \cite{local}                    & 1x &  128 & 200 & 58.8 \\
MoCo V2 \cite{mocov2}                    & 1x &  256 &  200 & 67.5 \\
MoCo V2 - WeakerAug \cite{mocov2}  & 1x &  256 &  200 & 68.3  \\
MoCHi \cite{mochi}                       & 1x &  512 &  200 & 68.0 \\
CPC v2 \cite{cpc}                        & 1x &  512 &  200 & 63.8 \\
PCL v2 \cite{PCL}                        & 1x &  256 &  200 & 67.6 \\
AdCo \cite{adco}                         & 1x &  256 &  200 & 68.6 \\
InfoMin \cite{goodview}                  & 1x &  256 &  200 & 70.1 \\
ISD \cite{ISD}                           & 1x &  256 &  200 & 69.8 \\
\textcolor{collapsed}{ReSSL * (1 Crop) \cite{ressl}}   & \textcolor{collapsed}{1x}  & \textcolor{collapsed}{256} &  \textcolor{collapsed}{200} & \textcolor{collapsed}{69.9} \\
\textbf{ReSSL (1 Crop)}                  & \textbf{1x}  & \textbf{1024} &  \textbf{200} & \textbf{72.0} \\
\bottomrule
\emph{2x Backprop Methods } \\
CLSA-Single \cite{stronger}              & 2x &  256  &  200 & 69.4 \\
SimCLR \cite{simclr}                     & 2x &  4096 &  200 & 66.8 \\
SimSiam \cite{SimSiam}  & 2x   & 256  &  200 & 70.0 \\
MoCo V2 - Symmetric \cite{SimSiam}   & 2x  & 256 & 200 & 69.9 \\ 
WCL \cite{WCL}   & 2x  & 4096 & 200 & 70.3 \\
NNCLR \cite{nnclr}   & 2x  & 256 & 200 & 70.7 \\ 
BYOL \cite{byol}       & 2x  & 4096 & 300 & 72.5 \\
Barlow Twins \cite{barlowtwins}       & 2x  & 2048 & 300 & 71.4 \\
MoCo V3 \cite{mocov3}  & 2x  & 4096 & 300 & 72.8 \\
SwAV \cite{swav}                         & 2x &  4096 &  400 & 70.1 \\
W-MSE 4 \cite{wmse}   & 2x & 1024 & 400 & 72.6 \\
\textcolor{collapsed}{ReSSL * (2 Crop) \cite{ressl}} & \textcolor{collapsed}{2x}  & \textcolor{collapsed}{256} &  \textcolor{collapsed}{200} & \textcolor{collapsed}{71.4} \\
\textbf{ReSSL (2 Crops)}   & \textbf{2x}  & \textbf{1024} &  \textbf{200} & \textbf{73.2} \\
\bottomrule
\end{tabular}
\vspace{-8pt}
\end{table}

\textbf{Working with Multi-Crops and Stronger Augmentations}.
We also performed ReSSL with multi-Crop strategy and stronger augmentations. Specifically, we follow the strategy from \cite{adco} which adopts 5 augmented views with the resolution of $224 \times 224, 192 \times 192, 160 \times 160, 128 \times 128, 96 \times 96 $, and set the min / max crop area to (0.14, 0.117, 0.095, 0.073, 0.05) / (1.0,  0.86,  0.715, 0.571, 0.429) for the 5 views respectively, the rest of augmentation strategy is consistent with our 1x and 2x setting. The result is shown in Table \ref{table:multi_crop}. Notably, with 200 epochs of training, ReSSL achieves 76.0\% Top-1 accuracy on ImageNet, which is significantly better than previous methods with even 800 or 1000 epochs of training. We tested our ReSSL model with extended training, specifically for 400 epochs, and found that it achieved the same performance as when trained for 200 epochs. This indicates that ReSSL converges more easily compared to other methods. Meanwhile, we also explored the effective of stronger augmentation to ReSSL. Concretely, we follow the same augmentation setting in \cite{stronger} and optimize ReSSL for 200 epoch. In this case, ReSSL achieves 76.3\% Top-1 accuracy which is almost on pair with the supervised baseline (76.5\%).

\renewcommand\arraystretch{1.0}
\begin{table}[h]
 \centering
 \setlength\tabcolsep{5pt}
 \vspace{-10pt}
 \caption{Working with Multi-Crops Strategy and Stronger Augmentations. (Linear Evaluation on ImageNet)}
 \vspace{-10pt}
 \label{table:multi_crop}
\begin{tabular}{l c c c}
\toprule 
Method & Epochs & Batch Size & Top-1\\ \hline
\emph{With Contrastive Augmentation} \\
SwAV \cite{swav} & 200 & 4096 & 72.7 \\
SwAV \cite{swav} & 800 & 4096 & 75.3 \\
AdCo \cite{adco} & 200 & 256  & 73.2 \\
AdCo \cite{adco} & 800 & 1024 & 75.7 \\
WCL \cite{WCL}   & 200 & 4096  & 73.3 \\
WCL \cite{WCL}   & 800 & 4096  & 74.7 \\
FCN \cite{FNCancel} & 1000 & 4096  & 74.4 \\
truncated triplet \cite{triplet} & 200 & 4160 & 74.1 \\
DINO \cite{dino} & 800 & 4096 & 75.3 \\
NNCLR \cite{nnclr} & 1000 & 4096 & 75.6 \\
UniGrad \cite{UniGrad} & 800 & 4096 & 75.5 \\ 
\textbf{ReSSL (Multi)} & \textbf{200} & \textbf{1024} & \textbf{76.0} \\ 
ReSSL (Multi) & 400 & 1024 & 76.0 \\ 
\bottomrule
\emph{With Stronger Augmentation} \\
CLSA-Multi \cite{stronger} & 200 & 256 & 73.3 \\
CLSA-Multi \cite{stronger} & 800 & 256 & 76.2 \\ 
\textbf{ReSSL (Strong)} & \textbf{200} & \textbf{1024} & \textbf{76.3} \\
\bottomrule
\end{tabular}
\vspace{-5pt}
\end{table}

\textbf{Working With Vision Transformer}
To underscore the broad applicability of our methodology, we additionally applied ReSSL to the Vision Transformer architecture \cite{vit, deit}. This included experiments with both the ViT-Small and ViT-Base models, with findings presented in Table \ref{table:vit}. The results, particularly under the 2 crops setting, highlight ReSSL's robust performance and its adaptability to various architectures, further attesting to its wide-ranging efficacy.

\renewcommand\arraystretch{1.0}
\begin{table}[h]
 \centering
 \setlength\tabcolsep{12pt}
 \vspace{-10pt}
 \caption{Performance on Vision Transformers. All methods use two 244 x 224 crops during training. The results of BYOL, SwAV, and SimCLR are copied from \cite{dino} and \cite{mocov3}. * denotes our reproduced result.}
 \vspace{-10pt}
 \label{table:vit}
\begin{tabular}{l c c c}
\toprule 
Method & Arch & Epochs & Top-1\\ \hline
BYOL \cite{mocov3} & ViT-S & 300 & 71.0 \\
SimCLR \cite{mocov3} & ViT-S & 300 & 69.0 \\
SwAV \cite{mocov3} & ViT-S & 300 & 67.1 \\ 
MoCo V3 \cite{mocov3} & ViT-S & 300 & 72.5 \\
BYOL \cite{dino} & ViT-S & 800 & 71.4 \\
MoCo V2 \cite{dino} & ViT-S & 800 & 72.7 \\
DINO \cite{dino} & ViT-S & 300 & 72.5 \\
\textbf{ReSSL (Ours)} & \textbf{ViT-S}  & \textbf{300} &  \textbf{73.8} \\ \midrule
BYOL \cite{mocov3} & ViT-B & 300 & 73.9 \\
SimCLR \cite{mocov3} & ViT-B & 300 & 73.9 \\
SwAV \cite{mocov3} & ViT-B & 300 & 71.6 \\ 
MoCo V3 \cite{mocov3} & ViT-B & 300 & 76.5 \\
DINO* \cite{dino} & ViT-B & 300 & 71.0 \\
\textbf{ReSSL (Ours)} & \textbf{ViT-B} & \textbf{300} & \textbf{76.8}  \\ \bottomrule
\end{tabular}
\vspace{-5pt}
\end{table}

\renewcommand\arraystretch{1.0}
\begin{table*}
    \centering
    \caption{Transfer Learning Performance: Linear evaluation and Fine-Tuning with ResNet-50 Pretrained on ImageNet. The performance of SimCLR, BYOL, and NNCLR is directly copied from their respective original papers. Following the evaluation protocol outlined in \cite{simclr}, we report Top-1 accuracy for all datasets except Pets, Flowers, and Caltech101, for which we report mean per-class accuracy. Our ReSSL method surpasses previous approaches on 6 out of the 9 datasets in both linear evaluation and fine-tuning settings.}
    \vspace{-10pt}
    \begin{tabular}{l c c c c c c c c c c c}
    \toprule
        Method & Epochs & CIFAR-10 & CIFAR-100 &  Food-101 & Cars & DTD & Pets & Flowers & Aircraft & Caltech101 &  Mean \\ \hline
        \emph{Linear Eval}\\
        Supervised & - & 93.6 & 78.3 & 72.3 & 66.7 & 74.9 & 91.5 & 94.7 & 61.0 & \textbf{94.5} & 80.8 \\
        SimCLR \cite{simclr} & 1000 & 90.5 & 74.4 & 72.8 & 49.3 & 75.7 & 84.6 & 92.6 & 50.3 & 89.3 & 75.5 \\
        BYOL \cite{byol} & 1000 & 91.3 & 78.4 & 75.3 & 67.8 & 75.5 & 90.4 & \textbf{96.1} & 60.6 & 94.2 & 81.1 \\
        NNCLR \cite{nnclr} & 1000 & 93.7 & \textbf{79.0} & 76.7 & 67.1 & 75.5 & \textbf{91.8} & 95.1 & 64.1 & 91.3 & 81.6 \\ 
        \textbf{ReSSL (Ours)} & 200 & \textbf{93.8} & 78.5 & \textbf{77.2} & \textbf{68.3} & \textbf{76.6} & 90.7 & 94.8 & \textbf{65.1} & \textbf{94.5} & \textbf{82.2} \\ \midrule

        \emph{Fine-tune} \\
        Supervised & - & 97.5 & \textbf{86.4} & 88.3 & \textbf{92.1} & 74.6 & 92.1 & 97.6 & 86.0 & 93.3 & 89.8 \\
        Random Init & - & 95.9 & 80.2 & 86.9 & 91.4 & 64.8 & 81.5 & 92.0 & 85.9 & 72.6 & 83.5 \\
        SimCLR & 1000 & 97.7 & 85.9 & 88.2 & 91.3 & 73.2 & 89.2 & 97.0 & 88.1 & 92.1 & 89.2 \\
        BYOL & 1000 & \textbf{97.8} & 86.1 & 88.5 & 91.6 & 76.2 & 91.7 & 97.0 & 88.1 & \textbf{93.8} & 90.1 \\
        \textbf{ReSSL (Ours)} & 200 & 97.7 & 85.9 & \textbf{88.6} & 91.8 & \textbf{77.3} & \textbf{93.0}  & \textbf{98.0} & \textbf{89.2} & \textbf{93.8} & \textbf{90.6} \\

        \bottomrule
    \end{tabular}
    \label{table:transfer_result}
    \vspace{-10pt}
\end{table*}

\textbf{Transfer learning on Linear Classification}.  
We show representations learned by ReSSL are effective for transfer learning on multiple downstream classification tasks. We follow the linear evaluation setup described in \cite{simclr, byol}. Specifically, we trained an L2-regularized multinomial logistic regression classifier on features extracted from the frozen pretrained network, then we used L-BFGS \cite{lbfgs} to optimize the softmax cross-entropy objective, we did not apply data augmentation. We selected the best L2-regularization parameter and learning rate from validation splits and applied it to the test sets. The datasets used in this benchmark are as follows: CIFAR-10 \cite{cifar}, CIFAR-100 \cite{cifar}, Food101 \cite{food101}, Cars \cite{cars}, DTD \cite{dtd}, Oxford-IIIT-Pets \cite{pets}, Aircrat \cite{aircraft}, Oxford-Flowers \cite{flowers}, and Caltech-101 \cite{caltech101}. We present the results in Table \ref{table:transfer_result}. It is clearly to see that ReSSL achieves state-of-the-art performance on CIFAR-10, Food-101, Cars, DTD, Aircraft, and Caltech101, which is significantly better than SimCLR, BYOL and NNCLR. we also assess the performance of ReSSL in the fine-tuning scenario, adhering strictly to the experimental setup outlined in previous works such as \cite{simclr, byol}. This involves employing random crops resized to 224 × 224 pixels and random flips during training. During testing, images are resized to 256 pixels along the shorter side, followed by extraction of a central crop sized 224 × 224 pixels. We conduct a grid search for optimal hyperparameters on a validation set, including the learning rate and weight decay. The learning rate ranges logarithmically from 0.0001 to 0.1, while weight decay ranges between $10^{-6}$ and $10^{-3}$ as well as 0, with the latter values being divided by the learning rate. The results, presented in the second section of Table \ref{table:transfer_result}, indicate that our ReSSL consistently outperforms prior methods on 6 datasets and achieves the highest mean accuracy across all 9 datasets.

\renewcommand\arraystretch{1.0}
\begin{table}[h]
 \centering
 \setlength\tabcolsep{5pt}
 \vspace{-5pt}
 \caption{ImageNet semi-supervised evaluation.}
 \vspace{-10pt}
 \label{table:semi}
\begin{tabular}{l  c  c c  c  c} 
\toprule 
 & & 1\% &  & 10\% &  \\
Method & Epochs & Top-1 & Top-5 & Top-1 & Top-5 \\
\hline
\emph{Semi-supervised} \\
S4L \cite{s4l} & 200 & - & 53.4 & - & 83.8 \\
UDA \cite{uda} & - & - & 68.8 & - & 88.5 \\
FixMatch \cite{fixmatch} & 300 & - & - & 71.5 & 89.1 \\
\hline
\emph{Self-Supervised} \\
NPID \cite{instance_discrimination} & 200 & - & 39.2 & - & 77.4 \\
PCL \cite{PCL} & 200 & - & 75.6 & - & 86.2 \\
PIRL \cite{pirl} & 800 & 30.7 & 60.4 & 57.2 & 83.8 \\
SwAV \cite{swav} & 800 & 53.9 & 78.5 & 70.2 & 89.9 \\ 
SimCLR \cite{simclr} & 1000 & 48.3 & 75.5 & 65.6 & 87.8 \\ 
BYOL \cite{byol} & 1000 & 53.2 & 78.4 & 68.8 & 89.0 \\ 
Barlow Twins \cite{barlowtwins} & 1000 & 55.0 & 79.2 & 69.7 & 89.3 \\
NNCLR \cite{nnclr} & 1000 & 56.4 & 80.7 & 69.8 & 89.3 \\
\textbf{ReSSL (Ours)} & \textbf{200} & \textbf{57.8} & \textbf{81.6} & \textbf{71.2} & \textbf{90.2} \\
\bottomrule
\end{tabular}
% \vspace{-5pt}
\end{table}

\textbf{Semi-Supervised Learning}. 
Next, we evaluate the performance obtained when fine-tuning the model representation using only 1\% and 10\% labeled examples. We directly adopt the labeled and unlabeled file list from \cite{simclr} for fair comparison. For 1\%, we freeze the backbone layers and set the learning rate to 0.1 for the final linear layer. For 10\%, we use a learning rate of 0.02 and 0.2 for the backbone layers and the final linear layer. The model will be optimized for 60 and 20 epochs for 1\% and 10\% setting with a cosine learning rate scheduler. We do not apply any weight decay in this experiments. The result has been shown in Table \ref{table:semi}. With only 200 epochs of training, ReSSL outperforms NNCLR 1.4\% and SwAV 1\% with 1\% and 10\% labeled examples, demonstrating the superiority of the representation quality learned by ReSSL.

\renewcommand\arraystretch{1.0}
\begin{table}[h]
 \centering
 \setlength\tabcolsep{10pt}
 \vspace{-5pt}
 \caption{Fine-tuning on Full ImageNet}
 \vspace{-10pt}
 \label{table:ft-full}
\begin{tabular}{l  c  c c} 
\toprule 
Method & Epochs & Top-1 & Top-5 \\ \hline
SimCLR \cite{simclr} & 1000 & 76.5 & 93.5 \\
SimCLRv2 \cite{simclrv2} & 800 & 76.3 & - \\
BYOL \cite{byol} & 1000 & 77.7 & 93.9 \\
\textbf{ReSSL (Ours)} & 200 &  \textbf{78.3} & \textbf{94.2} \\
\bottomrule
\end{tabular}
\end{table}

\textbf{Fine-tuning on Full ImageNet}. We also investigate the efficacy of fine-tuning on the complete ImageNet dataset. In this experiment, we set the learning rate at 0.04 for the backbone layers and 0.01 for the linear layer. The model is trained over 30 epochs, with no weight decay. The results are presented in Table \ref{table:ft-full}. Essentially, our ReSSL surpasses previous benchmarks by a margin of 0.6\%, thereby demonstrating its superior performance.

\renewcommand\arraystretch{1.0}
\begin{table}[h]
 \centering
 \setlength\tabcolsep{5pt}
 \vspace{-5pt}
 \caption{Transfer learning on low-shot image classification. We follow the implementation in \cite{PCL} and reproduce the result of BYOL and SwAV using the official checkpoints.}
 \vspace{-10pt}
 \label{table:low-shot}
\begin{tabular}{l  c  c c  c  c c c} 
\toprule 
Method & Epochs & k=4 & k=8 & k=16 & k=32 & k=64 & Full \\
\hline
Random     & -  & 10.1 & 10.4 & 10.8 & 11.3 & 11.9 & 12.4 \\
Supervised & 90 & 73.8 & 79.5 & 82.3 & 84.0 & 85.1 & 87.3 \\ \hline
MoCo V2 \cite{mocov2} & 200 & 64.9 & 72.5 & 76.1 & 79.2 & 81.5 & 84.6 \\
PCL V2 \cite{PCL} & 200 & 66.2 & 74.5 & 78.3 & 80.7 & 82.7 & 85.4 \\
SwAV \cite{swav}  & 800 & 64.0 & 73.0 & 78.7 & 82.3 & 84.9 & 88.1 \\
BYOL \cite{byol}  & 1000 & 64.8 & 73.4 & 78.3 & 81.7 & 83.9 & 87.0 \\
\textbf{ReSSL (Ours)}  & \textbf{200} & \textbf{66.8} & \textbf{75.3} & \textbf{80.5} & \textbf{83.8} & \textbf{86.0} & \textbf{88.6} \\
\bottomrule
\end{tabular}
% \vspace{-10pt}
\end{table}

\textbf{Transfer Learning on Low-shot Classification}. 
We further evaluate the quality of the learned representations by transferring them to the low-shot classification task. Following \cite{PCL}, we perform linear classification on the PASCAL VOC2007 dataset \cite{pascal-voc-2007}. Specifically, we resize all images to 256 pixels along the shorter side and taking a 224 × 224 center crop. Then, we train a linear SVM on top of corresponding global average pooled final representations. To study the transferability of the representations in few-shot scenarios, we vary the number of labeled examples k. and report the mAP. Table \ref{table:low-shot} shows the comparison between our method with previous works. We report the average performance over 5 runs (except for k=full). Notably, ReSSL consistently has a higher performance than other methods across different settings, and it also surpasses the supervised setting when k is greater than 64.

\renewcommand\arraystretch{1.0}
\begin{table}[h]
 \centering
 \setlength\tabcolsep{10pt}
 \vspace{-5pt}
 \caption{Transfer Learning on Object Detection and Instance Segmentation.}
 \vspace{-10pt}
 \label{table:detection}
\begin{tabular}{l  c  c} 
\toprule 
Method & $AP^{Box}$ & $AP^{Mask}$ \\
\hline
Random & 35.6 & 31.4 \\
Supervised & 40.0 & 34.7 \\  \hline
Relative-Loc \cite{patch_prediction} & 40.0 & 35.0 \\
Rotation-Pred \cite{rotation_selfsup} & 40.0 & 34.9 \\
NPID \cite{instance_discrimination} & 39.4 & 34.5 \\
\textbf{MoCo V2} \cite{mocov2} & \textbf{40.9}  & \textbf{35.5}  \\
SimCLR \cite{simclr} & 39.6 & 34.6 \\
BYOL \cite{byol} & 40.3 & 35.1 \\
\textbf{ReSSL (Ours)} & \textbf{40.4}  & \textbf{35.2} \\
\bottomrule
\end{tabular}
% \vspace{-5pt}
\end{table}

\textbf{Object Detection and Instance Segmentation}.
We finally evaluate the learned representations for the localization based tasks of object detection and instance segmentation on COCO \cite{coco} dataset by following the experiment setting in \cite{moco}. Specifically, we use the ReSSL pretrained weight to initialize the Mask R-CNN \cite{maskrcnn} C4 backbone, the model will be finetuned on the COCO 2017 \emph{train} split and report the results on the \emph{val} split. We use a learning rate of 0.04 and keep the other parameters the same as in the default 2x schedule in detectron2 \cite{detectron2}. The results in Table \ref{table:detection} demonstrates that our ReSSL has a competitive performance with the state-of-the-art contrastive learning methods for these localization tasks.

\renewcommand\arraystretch{0.8}
\begin{table*}
    \centering
    \caption{Transfer Learning Performance on Taskonomy. we report the L1 losses scaled by 1000. The lower number indicates the better performance. All methods employ pre-training with the ViT-Base on ImageNet, with the exception of CroCo \cite{croco}, which is trained on Habitat\cite{habitat19iccv}.}
    \vspace{-10pt}
    \begin{tabular}{l c c c c c c c c c }
    \toprule
        Method  & Curvature & Depth &  Edges & 2D-keypoints & 3D-keypoints & Normals & Occlusion & Reshading & Average  \\ \hline
        \emph{Contrastive Based} \\
        DINO \cite{dino} & 43.04 & 38.42 & 3.80 & 0.16 & 45.85 & 65.71 & 0.57 & 115.02 & 39.07 \\ 
        \textbf{ReSSL (Ours)} & 43.10 & 35.96 & 2.71 & 0.14 &  46.76  & 64.49 & \textbf{0.55} & 111.48 & 38.15 \\ \midrule
        \emph{MIM Based} \\
        MAE \cite{mae} & 41.59 & 35.83 & \textbf{1.19}  & 0.08 & 44.18 & 59.20 & \textbf{0.55} & 106.08 & 36.09 \\ 
        MultiMAE \cite{multimae} & 41.42 & 35.38  & 2.17 & \textbf{0.07} & 44.03 & 60.35 & 0.56 & 105.25 & 36.17  \\ 
        CroCo (Habitat) \cite{croco} & \textbf{40.91} & \textbf{31.34}  & 1.74 & 0.08 & \textbf{41.69} & \textbf{54.13} & \textbf{0.55} & \textbf{93.58} & \textbf{33.00} \\
        \bottomrule
    \end{tabular}
    \label{table:taskonomy}
    \vspace{-10pt}
\end{table*}

\renewcommand\arraystretch{0.8}
\begin{table}[h]
 \setlength\tabcolsep{5pt}
 \centering
 \caption{Top-1 accuracy under the linear evaluation on ImageNet with various lightweight backbones. The numbers for SimCLR, MoCo V2, SwAV, and BYOL on ResNet-18 are copied from \cite{oss}}
 \label{table:lightweight}
 \vspace{-10pt}
\begin{tabular}{l c c c c}
\toprule 
Method &  \makecell{Teacher\\Arch}  & \makecell{Teacher\\Top-1}  & Epochs & \makecell{Student\\Top-1} \\ \hline
\emph{ResNet-18} \\
SimCLR \cite{simclr} & - & - & 800 & 53.4 \\
MoCo V2 \cite{moco} & - & - & 800 & 56.1 \\
SwAV \cite{swav} & - & - & 800 & 64.9 \\
BYOL \cite{byol} & - & - & 800 & 61.9 \\
SEED \cite{seed} & ResNet-50 & 67.5 & 200 & 57.9 \\
DisCo \cite{disco} & ResNet-50 & 67.5 & 200 & 60.6 \\
BINGO \cite{bingo} & ResNet-50 & 67.5 & 200 & 61.4 \\
CompRess \cite{compress} & ResNet-50 & 71.1 & 200 & 62.6 \\
OSS \cite{oss} & ResNet-50 & 75.3 & 200 & 64.1 \\
SimDis-Off \cite{oss} & ResNet-50 & 74.3 & 300 & 65.2 \\ 
\textbf{ReSSL (1 Crop)}  & \textbf{-} & \textbf{-} & \textbf{200} & \textbf{62.4} \\ 
\textbf{ReSSL (2 Crops)} & \textbf{-} & \textbf{-} & \textbf{200} & \textbf{63.8} \\ 
\textbf{ReSSL (Multi)}   & \textbf{-} & \textbf{-} & \textbf{200} & \textbf{66.5} \\ 
\bottomrule
\emph{ResNet-34} \\
MoCo V2 \cite{mocov2} & - & - & 200 & 57.4 \\
SEED \cite{seed} & ResNet-50 & 67.5 & 200 & 58.5 \\
SEED \cite{seed} & ResNet-152 & 74.2 & 200 & 62.7 \\
DisCo \cite{disco} & ResNet-50 & 67.5 & 200 & 62.5 \\
DisCo \cite{disco} & ResNet-152 & 74.1 & 200 & 68.1 \\
BINGO \cite{bingo} & ResNet-50 & 67.5 & 200 & 63.5 \\
BINGO \cite{bingo} & ResNet-152 & 74.1 & 200 & 69.1 \\ 
\textbf{ReSSL (1 Crop)}  & \textbf{-} & \textbf{-} & \textbf{200} & \textbf{66.3} \\ 
\textbf{ReSSL (2 Crops)} & \textbf{-} & \textbf{-} & \textbf{200} & \textbf{68.4} \\ 
\textbf{ReSSL (Multi)}   & \textbf{-} & \textbf{-} & \textbf{200} & \textbf{71.0} \\ 
\bottomrule
\emph{EfficientNet-B0} \\
MoCo V2 \cite{mocov2} & - & - & 200 & 46.8 \\
SEED \cite{seed} & ResNet-50 & 67.5 & 200 & 61.3 \\
SEED \cite{seed} & ResNet-152 & 74.2 & 200 & 65.3 \\
DisCo \cite{disco} & ResNet-50 & 67.5 & 200 & 66.5 \\
DisCo \cite{disco} & ResNet-152 & 74.1 & 200 & 67.8 \\
OSS \cite{oss} & ResNet-50 & 75.3 & 200 & 64.1 \\ 
\textbf{ReSSL (1 Crop)}  & \textbf{-} & \textbf{-} & \textbf{200} & \textbf{69.7} \\ 
\textbf{ReSSL (2 Crops)} & \textbf{-} & \textbf{-} & \textbf{200} & \textbf{70.9} \\ 
\textbf{ReSSL (Multi)}   & \textbf{-} & \textbf{-} & \textbf{200} & \textbf{72.2} \\ 
\bottomrule
\emph{EfficientNet-B1} \\
MoCo V2 \cite{mocov2} & - & - & 200 & 48.4 \\
SEED \cite{seed} & ResNet-50 & 67.5 & 200 & 61.4 \\
SEED \cite{seed} & ResNet-152 & 74.2 & 200 & 67.3 \\
DisCo \cite{disco} & ResNet-50 & 67.5 & 200 & 66.6 \\
DisCo \cite{disco} & ResNet-152 & 74.1 & 200 & 73.1 \\ 
\textbf{ReSSL (1 Crop)}  & \textbf{-} & \textbf{-} & \textbf{200} & \textbf{71.1} \\ 
\textbf{ReSSL (2 Crops)} & \textbf{-} & \textbf{-} & \textbf{200} & \textbf{73.1} \\ 
\textbf{ReSSL (Multi)}   & \textbf{-} & \textbf{-} & \textbf{200} & \textbf{74.6} \\ 
\bottomrule
\emph{MobileNet-V3-Large} \\
MoCo V2 \cite{mocov2} & - & - & 200 & 36.2 \\
SEED \cite{seed} & ResNet-50 & 67.5 & 200 & 55.2 \\
SEED \cite{seed} & ResNet-152 & 74.2 & 200 & 61.4 \\
DisCo \cite{disco} & ResNet-50 & 67.5 & 200 & 64.4 \\
DisCo \cite{disco} & ResNet-152 & 74.1 & 200 & 63.7 \\ 
\textbf{ReSSL (1 Crop)}  & \textbf{-} & \textbf{-} & \textbf{200} & \textbf{64.7} \\
\textbf{ReSSL (2 Crops)} & \textbf{-} & \textbf{-} & \textbf{200} & \textbf{66.6} \\
\textbf{ReSSL (Multi)}   & \textbf{-} & \textbf{-} & \textbf{200} & \textbf{68.2} \\ 
\bottomrule
\vspace{-20pt}
\end{tabular}
\end{table}

\textbf{Transfer Learning on Taskonomy}.  Recently, numerous studies, including those by \cite{multimae, croco, croco_v2}, have begun to explore the transferability of pre-trained models for 3D vision tasks. To ensure the thoroughness of this research, we conducted a transfer of our pre-trained ViT-Base model (referenced in Table \ref{table:vit}) to the Taskonomy \cite{taskonomy} dataset, aiming to assess its performance on 3D vision tasks. We follow the same setting as in \cite{multimae}, where we use the taskonomy splits and fine-tune the model on a subset of 800 training images, we finally report L1 losses on the tiny-split test set, as reported in Table \ref{table:taskonomy}. Our findings indicate that while our ReSSL model outperforms DINO, but methods based on Masked Image Modeling (MIM) such as \cite{mae, multimae, croco} generally surpass contrastive-based approaches. This aligns with our intuition, given that contrastive-based methods like DINO, MoCo, and our ReSSL tend to focus on capturing the global semantics of an image, which is advantageous for tasks like classification. Conversely, MIM-based methods emphasize predicting masked patches or patch features, directing more attention to the interactions among local image patches. This focus is particularly beneficial for tasks requiring detailed geometric information. Integrating MIM-based methods with contrastive-based approaches could potentially enhance the transfer performance for 3D vision tasks. We would like to explore this idea in our future work.

\renewcommand\arraystretch{1.0}
\begin{table}[h]
 \setlength\tabcolsep{5pt}
 \centering
 \caption{Semi-supervised learning for lightweight architectures.}
 \label{table:lightweit_semi}
 \vspace{-10pt}
\begin{tabular}{l c c c c }
\toprule 
Method &  \makecell{Teacher\\Arch}  & \makecell{Teacher\\Top-1}  &  1\% & 10\% \\ \hline
\emph{ResNet-18} \\
SimCLR \cite{simclr} & - & - & 28.8 & 54.2 \\
MoCo V2 \cite{mocov2} & - & - & 25.2 & 54.1 \\
BYOL \cite{byol} & - & - & 25.6 & 52.3 \\
SwAV \cite{swav} & - & - & 39.7 & 60.4 \\
Compress \cite{compress} & ResNet-50 & 67.5 & 41.2 & 47.6 \\
SEED \cite{seed} & ResNet-50 & 67.5 & 39.1 & 50.2 \\
DisCo \cite{disco} & ResNet-50 & 67.5 & 39.2 & 50.1 \\
BINGO \cite{disco} & ResNet-50 & 67.5 & 42.8 & 57.5 \\
\textbf{ReSSL (Multi)}   & \textbf{-} & \textbf{-} & \textbf{45.6} & \textbf{62.3} \\ \bottomrule
\emph{ResNet-34} \\
\textbf{ReSSL (Multi)}   & \textbf{-} & \textbf{-} & \textbf{55.4} & \textbf{67.8} \\ \bottomrule
\emph{EfficientNet-B0} \\
SEED \cite{seed} & ResNet-50 & 67.5 & 37.7 & 53.3 \\
SEED \cite{seed} & ResNet-152 & 74.1 & 41.8 & 56.7 \\
DisCo \cite{disco} & ResNet-50 & 67.5 & 42.1 & 56.6 \\
DisCo \cite{disco} & ResNet-152 & 74.1 & 52.0 & 63.1 \\
\textbf{ReSSL (Multi)}   & \textbf{-} & \textbf{-} & \textbf{55.6} & \textbf{68.2} \\ 
\bottomrule
\emph{EfficientNet-B1} \\
\textbf{ReSSL (Multi)}   & \textbf{-} & \textbf{-} & \textbf{59.5} & \textbf{70.9} \\ \bottomrule
\emph{MobileNet-V3-Large} \\
SEED \cite{seed} & ResNet-50 & 67.5 & 36.0 & 51.7 \\
SEED \cite{seed} & ResNet-152 & 74.1 & 40.0 & 52.8 \\
DisCo \cite{disco} & ResNet-50 & 67.5 & 38.8 & 51.3 \\
DisCo \cite{disco} & ResNet-152 & 74.1 & 42.9 & 54.4 \\
\textbf{ReSSL (Multi)}   & \textbf{-} & \textbf{-} & \textbf{50.9} & \textbf{62.8} \\ 
\bottomrule
\vspace{-10pt}
\end{tabular}
\end{table}

\subsection{Results on Lightweight Architectures}
We also applied our ReSSL on various lightweight architectures to further demonstrate the generality. By following the experimental setting in \cite{seed, disco}, we performed ReSSL on ResNet-18, ResNet-34 \cite{resnet}, EfficientNet-B0, EfficientNet-B1 \cite{efficientnet}, and MobileNetV3-Large \cite{mbv3}. The linear evaluation result has been shown in  Table \ref{table:lightweight}. We first present the result for SimCLR, MoCo V2, BYOL and SwAV on ResNet-18. Obviously, these methods have significant gaps with the supervised performance even with 800 epochs of training. Although the Knowledge distillation (KD) based methods (\eg SEED, Compress, DisCo and BINGO) can somewhat improve the performance, our ReSSL is much more efficient since it does not require a large teacher. More importantly, the performance of ReSSL is consistently better than these KD methods especially on those extremely small networks (\eg EfficientNet-B0 and Mobilenet-V3-Large). 

Apart from the linear evaluation, we also report the performance for semi-supervised learning. We follow the same training strategy as our ResNet-50 experiments. Table \ref{table:lightweit_semi} shows that our ReSSL achieves promising results for various lightweight architectures and outperforms the latest self-supervised KD methods by a clear margin.

\subsection{More Ablation Studies on ImageNet}
In this section, we delve deeper with extra ablation studies to show the impact that the InfoNCE Warm-up and Predictor strategy have on ImageNet performance. Additionally, we explore the intriguing properties similar to those discussed in \cite{dino}. For these experiments, we simplify the process by training our model over 100 epochs using a single-crop setting.

\renewcommand\arraystretch{1.0}
\begin{table}[h]
 \centering
 \caption{Effect of InfoNCE Warm-up and Predictor}
 \vspace{-10pt}
 \label{table:in_ablation}
\begin{tabular}{c c c c } 
\toprule 
InfoNCE Warm-up & Predictor & Top-1 & Improvements \\ \hline
& & 68.0 & -  \\
$\checkmark$ & & 68.4 & +0.4  \\
 & $\checkmark$ & 69.6 & +1.6 \\
$\checkmark$ & $\checkmark$ & 70.3 & +2.3 \\
\toprule 
\end{tabular}
% \vspace{-10pt}
\end{table}

\textbf{Effect of InfoNCE Warm-up and Predictor}
We present the impact of InfoNCE Warm-up and Predictor on ImageNet experiments in Table \ref{table:in_ablation}. It is observed that these two strategies enhance the performance of ReSSL, elevating it from 68.0\% to 70.3\%, which underscores their effectiveness.

\renewcommand\arraystretch{1.0}
\begin{table}[h]
 \centering
 \caption{Effect of Temperature Warm-up}
 \vspace{-10pt}
 \label{table:temp_warmup}
\begin{tabular}{c c c c}
\toprule 
 & Baseline & 0.03 $\rightarrow$ 0.06 & 0.04 $\rightarrow$ 0.07  \\ \hline
Top-1 & 70.3 & 69.4 & 69.7 \\
\toprule 
\end{tabular}
% \vspace{-10pt}
\end{table}

\textbf{Effect of Temperature Warm-up}
Dino proposed an innovative warm-up strategy for temperature, starting with a lower temperature for the teacher logits to avert model collapse, and then gradually increasing it. This method showed a slight improvement. To evaluate this approach, we conducted two experiments, as outlined in Table \ref{table:temp_warmup}. However, unlike DINO, this strategy negatively impacted the performance of ReSSL, leading us to not incorporate it in our research.

\renewcommand\arraystretch{1.0}
\begin{table}[h]
 \centering
 \setlength\tabcolsep{15pt}
 \caption{Teacher Performance vs. Student Performance}
 \vspace{-10pt}
 \label{table:student_teacher}
\begin{tabular}{c c c }
\toprule 
  & Student & Teacher \\ \hline
ResNet-50 & 70.3 & 69.8 \\
ViT-Small & 73.8 &  73.8 \\
\bottomrule 
\end{tabular}
% \vspace{-10pt}
\end{table}

\textbf{Teacher Performance vs. Student Performance}
We also examine the performance of the teacher and student models, with the outcomes detailed in Table \ref{table:student_teacher}. Showing that the student model outperforms the teacher model when using ResNet-50. This discrepancy arises from the different augmentations applied to the teacher and student models, which leads to variations in Batch Norm statistics. For the teacher model, the parameter is optimized with strong augmentation, while its Batch Norm statistics are calculated using weak augmentation, resulting in performance disparities. To further investigate this phenomenon, we also evaluated the teacher and student models using the ViT-Small architecture, which substitutes the Batch Norm layer with a Layer Norm layer. This modification addresses the observed discrepancy, resulting in equivalent performance between the teacher and student models. Consequently, the student model will serve as the standard configuration for this research to ensure consistency.
\section{Conclusion}

In this work, we propose relational self-supervised learning (ReSSL), a new paradigm for unsupervised visual representation learning framework that maintains the relational consistency between instances under different augmentations. Our proposed ReSSL relaxes the typical constraints in contrastive learning where different instances do not always need to be pushed away on the embedding space, and the augmented views do not need to share exactly the same feature. An extensive empirical study shows the effect of each component in our framework. The experiments on large-scaled datasets demonstrate the efficiency and state-of-the-art performance for unsupervised representation learning.

% \appendices

% % you can choose not to have a title for an appendix
% % if you want by leaving the argument blank
% \section{}

% \vspace{-10pt}

% \ifCLASSOPTIONcompsoc
%   % The Computer Society usually uses the plural form
%   \section*{Acknowledgments}
% \else
%   % regular IEEE prefers the singular form
% \fi

% This work is funded by the National Key Research and Development Program of China (No. 2018AAA0100701) and the NSFC 61876095. Chang Xu was supported in part by the Australian Research Council under Projects DE180101438 and DP210101859. 

% \vspace{-10pt}

\ifCLASSOPTIONcaptionsoff
  \newpage
\fi



%\bibliographystyle{IEEEtran}
% \bibliography{IEEEabrv,../bib/paper}


% H.~Kopka and P.~W. Daly, \emph{A Guide to {\LaTeX}}, 3rd~ed.\hskip 1em plus

% \end{thebibliography}
% \bibliography{egbib}
% {\small
\bibliographystyle{plain}
\bibliography{egbib}
% }

\begin{IEEEbiography}[{\includegraphics[width=1in,height=1.25in,clip,keepaspectratio]{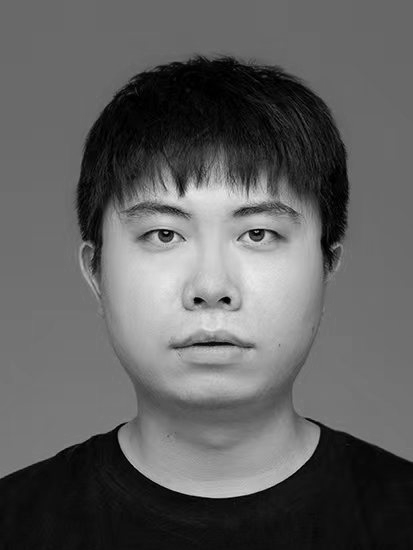}}]{Mingkai Zheng} received his BSc degree in computer science and technology from the University of Sydney (USYD), Australia, in 2019, where he is currently pursuing the Ph.D. degree. His research interests include pattern recognition and machine learning fundamentals with a focus on contrastive learning, self-supervised learning, unsupervised learning, and semi-supervised learning.
\end{IEEEbiography}
\begin{IEEEbiography}[{\includegraphics[width=1in,height=1.25in,clip,keepaspectratio]{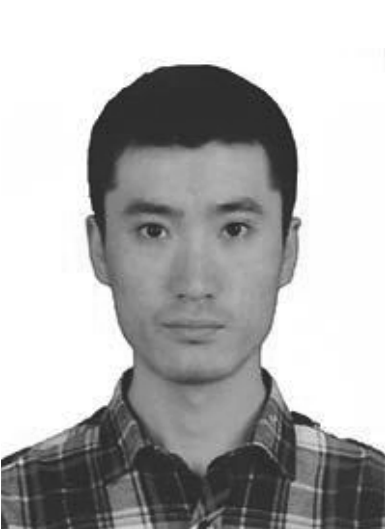}}]{Shan You} is currently a Senior Research Manager at SenseTime. Before that, he finished a post doc at Tsinghua University, and received a Ph.D. degree of computer science from Peking University. His research interests include fundamental algorithms for machine learning and computer vision, such as multimodal learning, AutoML, representation learning, and human-centric learning. He has published his research outcomes in many top tier conferences and transactions.
\end{IEEEbiography}
\begin{IEEEbiography}[{\includegraphics[width=1in,height=1.25in,clip,keepaspectratio]{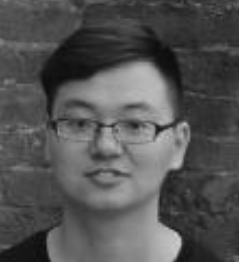}}]{Fei Wang} is the Director of SenseTime Intelligent Automotive Group. He is the head of SenseAuto-Parking engineering and SenseAuto-Cabin research. He leads a vibrant team of 60+ people to develop comprehensive solutions for the intelligent vehicle and deliver 20+ mass production of SenseAuto-Cabin projects in the last 6 years. He has published 20+ papers at CVPR/NIPS/ICCV during the last few years. Fei obtained his Bachelor’s degree and Master's degree from Beijing University of Posts and Telecommunications. Currently, he is a Ph.D. student at the University of Science and Technology of China. His research interests include Automotive Drive System, AI Chip, Deep Learning, etc.
\end{IEEEbiography}
\begin{IEEEbiography}[{\includegraphics[width=1in,height=1.25in,clip,keepaspectratio]{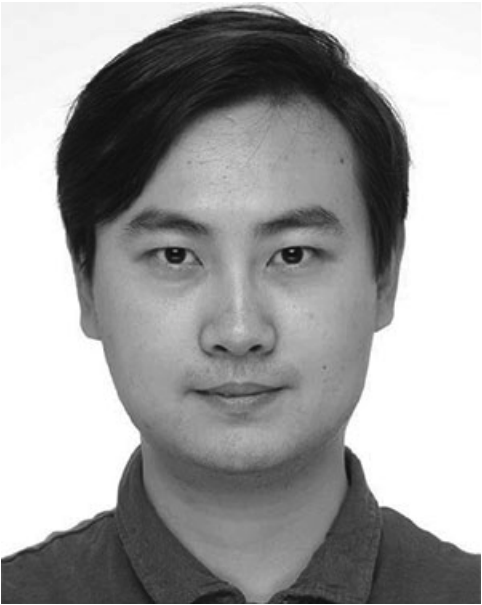}}]{Chen Qian} received the BEng degree from the Institute for Interdisciplinary Information Science, Tsinghua University, in 2012, and the MPhil degree from the Department of Information Engineering, the Chinese University of Hong Kong, in 2014. He is currently working at SenseTime as research director. His research interests include human-related computer vision and machine learning problems.
\end{IEEEbiography}
\begin{IEEEbiography}[{\includegraphics[width=1in,height=1.25in,clip,keepaspectratio]{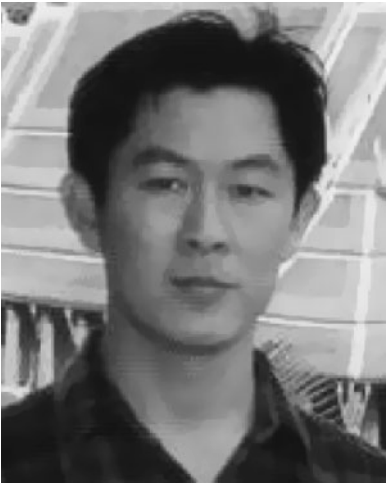}}]{Changshui Zhang} received the BSc degree in mathematics from Peking University, Beijing, China, in 1986, and the PhD degree from Tsinghua University, Beijing, China, in 1992. In 1992, he joined the Department of Automation, Tsinghua University, where he is currently a professor. His interests include pattern recognition, machine learning, etc. He has authored more than 200 papers. He currently serves on the editorial board of the journal Pattern Recognition. He is a member of the IEEE.
\end{IEEEbiography}
\begin{IEEEbiography}[{\includegraphics[width=1in,height=1.25in,clip,keepaspectratio]{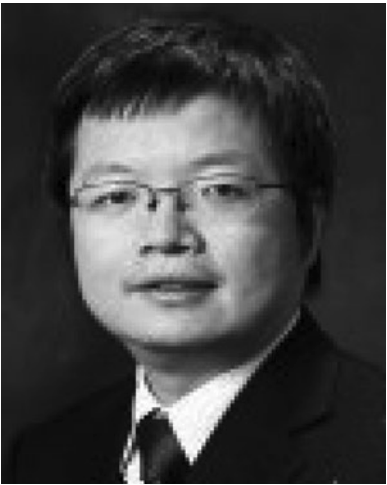}}]{Xiaogang Wang} received the BS degree from the University of Science and Technology of China, in 2001, the MS degree from The Chinese University of Hong Kong, in 2003, and the PhD degree from the Computer Science and Artificial Intelligence Laboratory, Massachusetts Institute of Technology, in 2009. He is currently an associate professor with the Department of Electronic Engineering, The Chinese University of Hong Kong. His research interests include computer vision and machine learning.
\end{IEEEbiography}
\begin{IEEEbiography}[{\includegraphics[width=1in,height=1.25in,clip,keepaspectratio]{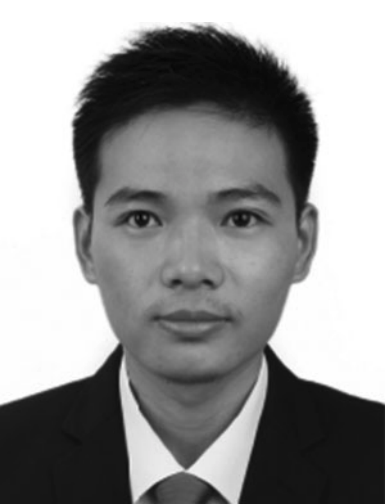}}]{Chang Xu} is an Associate Professor and ARC Future Fellow at the School of Computer Science, University of Sydney. He received the Ph.D. degree from Peking University, China. His research interests lie in machine learning algorithms and related applications in computer vision. He has published over 200 papers in prestigious journals and top tier conferences. He has received several paper awards, including Distinguished Paper Award in AAAI 2023 and Distinguished Paper Award in IJCAI 2018. He served as an area chair of NeurIPS, ICML, ICLR, KDD, CVPR, and MM, as well as a Senior PC member of AAAI and IJCAI. In addition, he served as an associate editor at IEEE T-PAMI, IEEE T-MM, and T-MLR. He has been named a Top Ten Distinguished Senior PC Member in IJCAI 2017 and an Outstanding Associate Editor at IEEE T-MM in 2022.
\end{IEEEbiography}

% \end{IEEEbiography}


% % insert where needed to balance the two columns on the last page with
% %\newpage



% \vfill

% \enlargethispage{-5in}

\end{document}